\documentclass{article}
\usepackage{PRIMEarxiv}
\usepackage[utf8]{inputenc} 
\usepackage[T1]{fontenc}    
\usepackage{hyperref}       
\usepackage{url}            
\usepackage{booktabs}       
\usepackage{amsfonts}       
\usepackage{nicefrac}       
\usepackage{microtype}      
\usepackage{lipsum}
\usepackage{fancyhdr}       
\usepackage{graphicx}       
\graphicspath{{media/}}     
\usepackage{subcaption}
\usepackage{multirow} 
\usepackage{amsmath}
\usepackage{diagbox}
\usepackage{siunitx}
\usepackage{comment}
\usepackage{nomencl}
\usepackage{makecell}
\usepackage[table]{xcolor}
\usepackage{array}
\makenomenclature

\newcolumntype{C}[1]{>{\centering\arraybackslash}m{#1}}
\usepackage{nomencl}
\title{Neural Field Ensembles for Aerodynamic Surface Prediction: Winning Solution to the ONERA CRM Wall Distribution 2025 Challenge

}
\author{
  \textbf{Lionel Salesses}\\
  Cenaero, Gosselies, Belgium \\
  \texttt{lionel.salesses@cenaero.be} \\
  \and
  \textbf{Caroline Sainvitu}\\
  Cenaero, Gosselies, Belgium \\
  \and
  \textbf{Tariq Benamara} \\
  Cenaero, Gosselies, Belgium \\
}
\begin{document}
\maketitle
\begin{abstract}
Machine-learning surrogate models offer a promising alternative to high-fidelity Computational Fluid Dynamics (CFD) simulations for aerodynamic analysis and design. However, constructing accurate surrogates for realistic aircraft configurations remain challenging due to complex geometries, multiple flow regimes, and limited training data.
This work presents the methodology that achieved first place in the ONERA CRM Wall Distribution Regression Challenge, which focuses on predicting pressure and skin-friction coefficient distributions over the NASA Common Research Model wing-body-pylon-nacelle configuration under different operating conditions.
The proposed approach formulates the problem as a conditional neural field mapping spatial coordinates, surface normals, and operating conditions to aerodynamic wall quantities.
Fourier feature encoding, a relative squared error objective aligned with the challenge metric, ensemble learning, and $k$-fold cross-validation are progressively introduced to improve prediction accuracy and exploit the limited training data.
Beyond presenting the final methodology, the paper documents the successive model design choices that led to the winning solution through a comprehensive ablation study and discusses several alternative approaches that were investigated but ultimately discarded. On the hidden competition test set, the proposed methodology achieves an overall score of 8.81, outperforming the strongest organizer-provided baseline, which achieved a score of 8.64, while requiring approximately three orders of magnitude fewer trainable parameters. These results illustrate that carefully designed coordinate-based neural fields constitute an efficient and robust framework for aerodynamic surrogate modeling on complex geometries under limited-data conditions.
\end{abstract}

\section{Introduction}
High-fidelity CFD simulations based on Reynolds-Averaged Navier–Stokes (RANS) equations are essential in aerodynamic design but remain computationally prohibitive for many-queries situations. This limitation has motivated the development of surrogate models capable of approximating flow solutions at a reduced computational expense. Despite recent advances in machine learning for scientific computing, constructing accurate surrogates for aerodynamic surface quantities remains challenging. The difficulty arises from the combination of complex geometries, highly nonlinear flow physics, and limited availability of high-fidelity data. In particular, transonic regimes introduce localized discontinuities associated with shock waves, while high angles of attack lead to flow separation and increased sensitivity to small variations in operating conditions. These phenomena make the construction of robust and accurate surrogate models particularly challenging.

The ONERA CRM wall distribution challenge provides a representative benchmark for aforementioned difficulties. The task consists in predicting pressure and friction fields over a realistic aircraft configuration for a wide range of aerodynamic conditions. The large number of surface points (260,774), the irregular nature of the geometry, and the diversity of flow regimes make conventional grid-based machine learning approaches difficult to apply directly. At the same time, the relatively limited number of available simulations imposes strong constraints on model complexity and generalization.

In this work, we present the methodology that achieved the best performance on the final hidden test set of the challenge. Our approach relies on coordinate-based neural field representations, which model aerodynamic quantities as continuous functions of spatial coordinates and flow parameters. This formulation naturally accommodates irregular geometries without requiring mesh connectivity information. To improve the representation of high-frequency flow structures, the neural fields are augmented with Fourier Feature Encoding (FFE) that mitigate the spectral bias of standard Multi-Layer Perceptrons (MLP). We further show that aligning the training objective with the challenge evaluation metric is critical for achieving robust performance, particularly on the most difficult flow configurations. Finally, we demonstrate that ensemble learning combined with $k$-fold cross-validation provides an effective strategy for improving generalization in a low-data regime.

\begin{figure}
    \centering
    \includegraphics[width=0.55\linewidth]{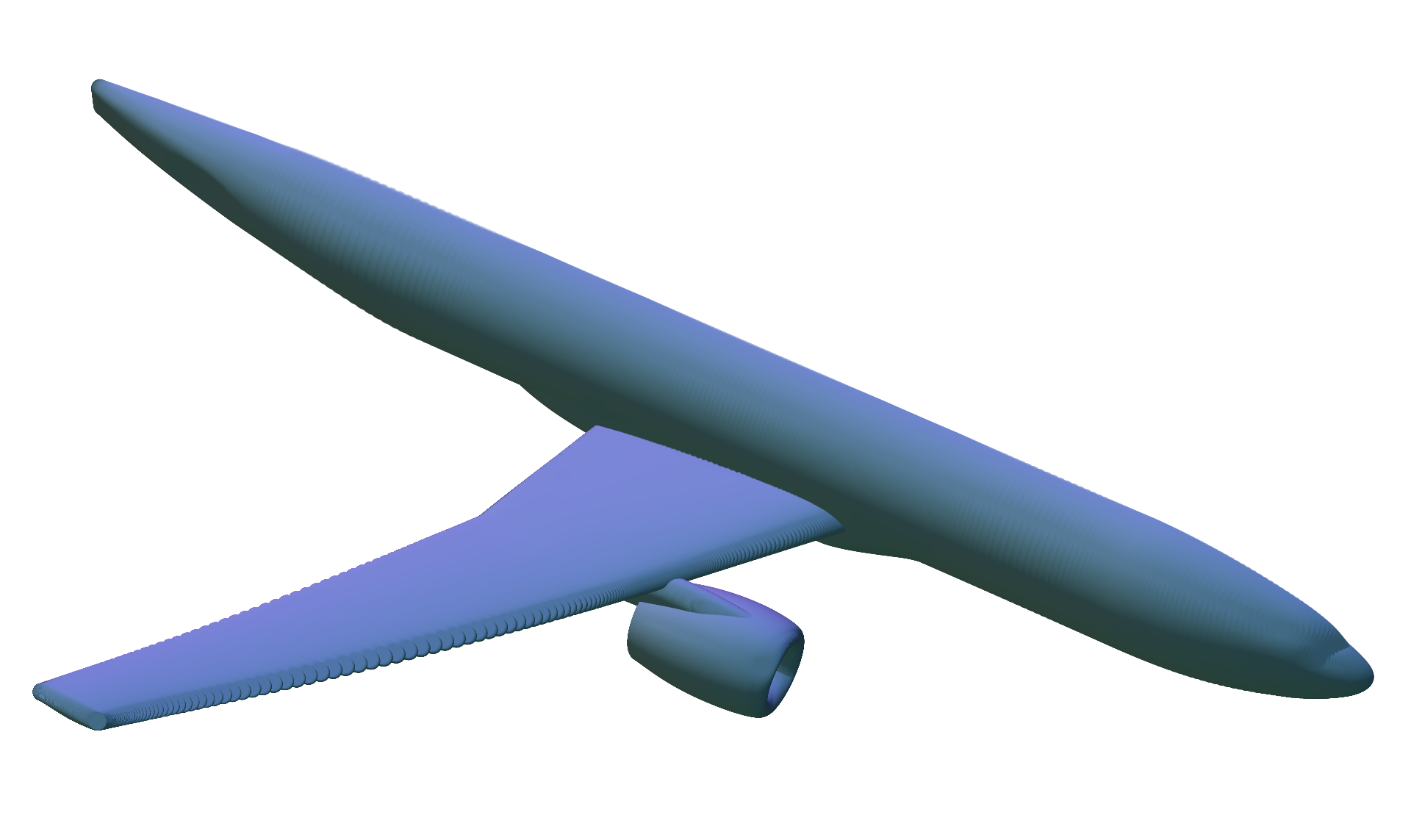}
    \caption{NASA Common Research Model geometry}
    \label{fig:geometry}
\end{figure}

\paragraph{Challenge context.}
This work originates from the ONERA CRM Wall Distribution Regression Challenge, organized by Jacques Peter and Quentin Bennehard (ONERA) and hosted on the Codabench platform\footnote{https://www.codabench.org/competitions/7535/}. The challenge was launched in May 2025 with the objective of evaluating machine learning approaches for the prediction of aerodynamic wall quantities on a realistic aircraft configuration representative of industrial applications. The benchmark is based on the CRM-WBPN database, introduced in Peter et al.~\cite{peter2025crm}, containing 468 RANS simulations performed on the NASA/Boeing Common Research Model Wing-Body-Pylon-Nacelle (CRM-WBPN) geometry (see Figure \ref{fig:geometry}) over a wide range of Mach numbers, Reynolds numbers and Angles of Attack (AoA). Participants were tasked with predicting wall distributions of pressure and friction coefficients from geometric descriptors and aerodynamic operating conditions. 

Beyond its competitive aspect, the challenge provided a common evaluation framework for comparing a broad spectrum of surrogate modeling strategies on a challenging aerodynamic regression problem. It attracted 64 participants from both academia and industry and received a total of 75 submissions. Following the completion of the challenge, the CRM-WBPN database was publicly released, making it possible to reproduce and extend the results obtained during the competition\footnote{https://entrepot.recherche.data.gouv.fr/dataset.xhtml?persistentId=doi:10.57745/K1UK6Z}.

Table~\ref{tab:leaderboard} summarizes the top-ranked solutions on the final hidden test set. The methodology presented in this paper achieved the best overall score among all submitted approaches. Beyond reporting the final solution, the objective of this paper is to analyze the successive methodological choices that led to this performance and to discuss the broader lessons that can be drawn for aerodynamic surrogate modeling using coordinate-based neural representations.
\begin{table}[ht]
    \centering
    \caption{Top-ranked solutions on the final hidden test set of the ONERA CRM Wall Distribution Regression Challenge, as reported on the leaderboard\protect\footnotemark.}  %
    \label{tab:leaderboard}
    \begin{tabular}{lllc}
    \toprule
    Rank & Team / Participant & Method & Final Score \\
    \midrule
    1 & Cenaero (this work) & Neural field ensembles & \textbf{8.8141} \\
    2 & ONERA baseline & Global MLP & 8.6401 \\
    3 & Air Liquide & Zonal MLP & 8.4764 \\
    4 & INTA & VAE + Gaussian process regression & 8.4370 \\
    5 & Centrale Supélec - Transvalor &  Physics informed operator  & 8.4254 \\
    \bottomrule
    \end{tabular}
\end{table}


Following the completion of the competition, the CRM-WBPN benchmark has also been used to evaluate subsequent surrogate modeling approaches. In particular, Ramos et al.~\cite{ramos2026fluidflow} introduced FluidFlow, a generative model based on conditional flow matching, and evaluated it on the same aircraft configuration. Published in 2026, this work did not participate in the original challenge but provides an independent reference for assessing more recent developments in aerodynamic surrogate modeling. Its results are therefore considered separately from the official competition ranking and are discussed in Section~\ref{sec:ff_results}.

\paragraph{Paper organization.}
The remainder of this paper is organized as follows. Section~\ref{section:related-work} reviews the most relevant literature on machine-learning surrogate models for computational fluid dynamics. Section~\ref{sec:problem_dataset} presents the CRM-WBPN benchmark and the evaluation protocol, while Section~\ref{sec:baselines} reviews the baseline models provided by the challenge organizers. Section~\ref{sec:methodology} describes the proposed neural-field formulation together with the main components of the final methodology. Section~\ref{sec:ablation} presents the progressive development of the winning solution through a comprehensive ablation study, evaluates the contribution of each methodological component, and concludes with a discussion of alternative approaches that were explored but ultimately discarded. Section~\ref{section:results} presents the final quantitative and qualitative results on the hidden test set. Finally, Section~\ref{section:conclusion} concludes the paper and outlines several directions for future research. 

\footnotetext[\thefootnote]{https://www.codabench.org/competitions/7535/\#/results-tab}

\section{Related Work} \label{section:related-work}
Machine-learning surrogate models have become an active research area for accelerating Computational Fluid Dynamics (CFD) simulations. Their objective is to approximate the mapping between geometric and aerodynamic parameters and the corresponding flow solution while reducing the computational cost by several orders of magnitude. Recent approaches can be broadly categorized into global regression methods, topology-aware neural networks, neural operators, and coordinate-based neural fields.

Early surrogate models relied on classical regression techniques, MLPs, or Convolutional Neural Networks (CNNs) to directly predict aerodynamic quantities from operating conditions. While these approaches can provide accurate predictions for restricted parameter spaces, their ability to represent complex nonlinear flow phenomena on realistic geometries remains limited \cite{bhatnagar2019prediction}, particularly when high-dimensional output fields must be predicted.

More recently, Graph Neural Networks (GNNs) have emerged as a natural framework for learning CFD solutions on unstructured meshes by explicitly exploiting mesh connectivity. Architectures such as MeshGraphNets demonstrated accurate prediction of transient fluid dynamics and solid mechanics by performing message passing over mesh elements~\cite{pfaff2020meshgraphnet}. Several subsequent works have extended graph-based operators to aerodynamic surrogate modeling on complex geometries and irregular meshes~\cite{li2023geometry}. Although these methods effectively exploit mesh topology, they require access to mesh connectivity, which is unavailable in the CRM-WBPN benchmark considered in this work.

In parallel, neural operators have been introduced to directly learn mappings between function spaces rather than finite-dimensional vectors. Fourier Neural Operators (FNO) demonstrated that entire families of PDE solutions can be approximated efficiently through spectral convolutions~\cite{li2020fourier}, while subsequent developments such as Geo-FNO extended these ideas to irregular geometries through learned domain deformations~\cite{li2023fourier}. Neural operators have become one of the main paradigms for data-driven PDE solving and surrogate modeling.

Another line of research relies on Implicit Neural Representations (INRs), also known as neural fields, which represent physical quantities as continuous functions of spatial coordinates. Originally developed for representing images, shapes and radiance fields, neural fields have subsequently been extended to PDE surrogate modeling and operator learning~\cite{serrano2023operator,kovachki2023neural}. Their continuous and discretization-independent formulation makes them particularly attractive for complex geometries and point-cloud representations. Recent advances, including Fourier feature encoding~\cite{tancik2020fourier} and sinusoidal representation networks (SIREN)~\cite{sitzmann2020implicit}, have substantially improved their ability to capture high-frequency signals and localized phenomena.

Neural fields have recently demonstrated excellent performance for aerodynamic surrogate modeling. Catalani et al.~\cite{catalani2024neural} proposed an INR framework capable of predicting aerodynamic fields on unstructured meshes while generalizing across unseen geometries. Their results showed that neural fields can outperform state-of-the-art GNNs on several aerodynamic benchmarks while remaining discretization independent. More recently, the MARIO framework further demonstrated the scalability of neural-field surrogates to large industrial aerodynamic datasets through efficient geometric conditioning and multiscale representations~\cite{catalani2025towards}.


Recent developments have also explored generative modeling as an alternative to conventional deterministic regression approaches. Ramos et al.~\cite{ramos2026fluidflow} introduced FluidFlow, a conditional flow-matching framework based on diffusion transformers for predicting aerodynamic fields directly on unstructured meshes. Although this work was published after the ONERA CRM Wall Distribution Regression Challenge and did not participate in the competition, the authors subsequently evaluated their approach on the same ONERA CRM-WBPN benchmark. Their results, which improve upon the organizer-provided Global MLP baseline, demonstrate the potential of generative models for aerodynamic surrogate modeling on complex geometries.

The methodology proposed in this work belongs to the family of conditional neural fields. Unlike previous studies focusing on multiple geometries or complete CFD meshes, we consider the challenge of predicting aerodynamic wall quantities on a fixed but complex aircraft geometry for a limited number of operating conditions and without access to mesh connectivity. The proposed methodology therefore focuses on carefully designing the model formulation and training pipeline, rather than introducing increasingly sophisticated neural architectures. In this respect, it provides a complementary, deterministic alternative to generative approaches such as FluidFlow.

\section{Problem Definition and Dataset} \label{sec:problem_dataset}
\subsection{Problem Definition}
\begin{figure}[htbp]
    \centering
    \begin{subfigure}[b]{0.32\textwidth}
        \centering
        \includegraphics[width=\textwidth]{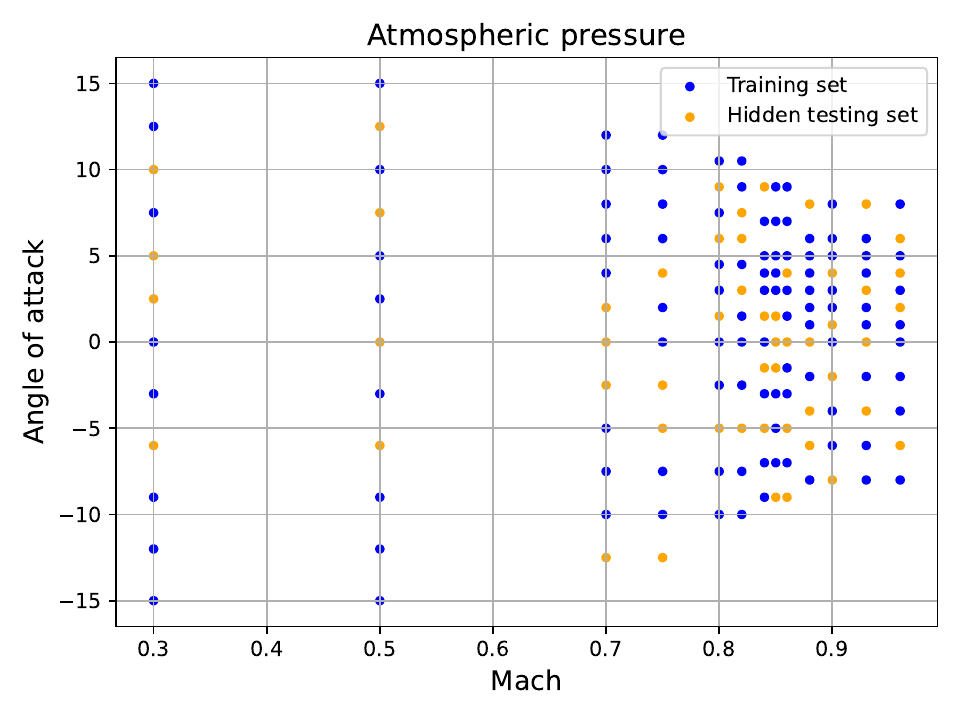}
        \label{fig:doe_low_pi}
    \end{subfigure}
    \hfill
    \begin{subfigure}[b]{0.32\textwidth}
        \centering
        \includegraphics[width=\textwidth]{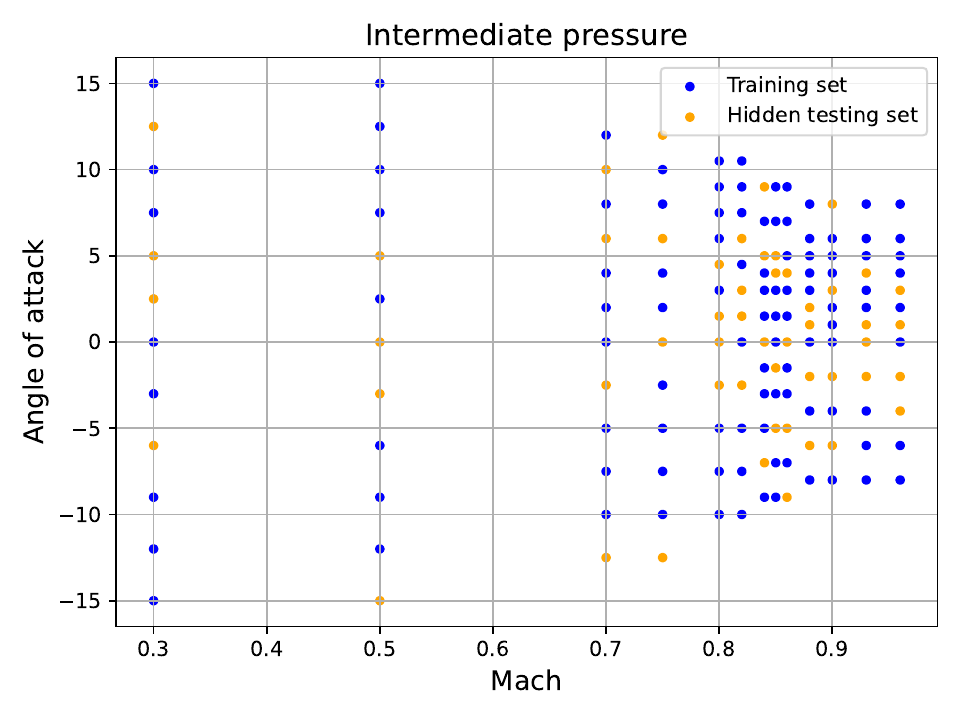}
        \label{fig:doe_mid_pi}
    \end{subfigure}
    \hfill
    \begin{subfigure}[b]{0.32\textwidth}
        \centering
        \includegraphics[width=\textwidth]{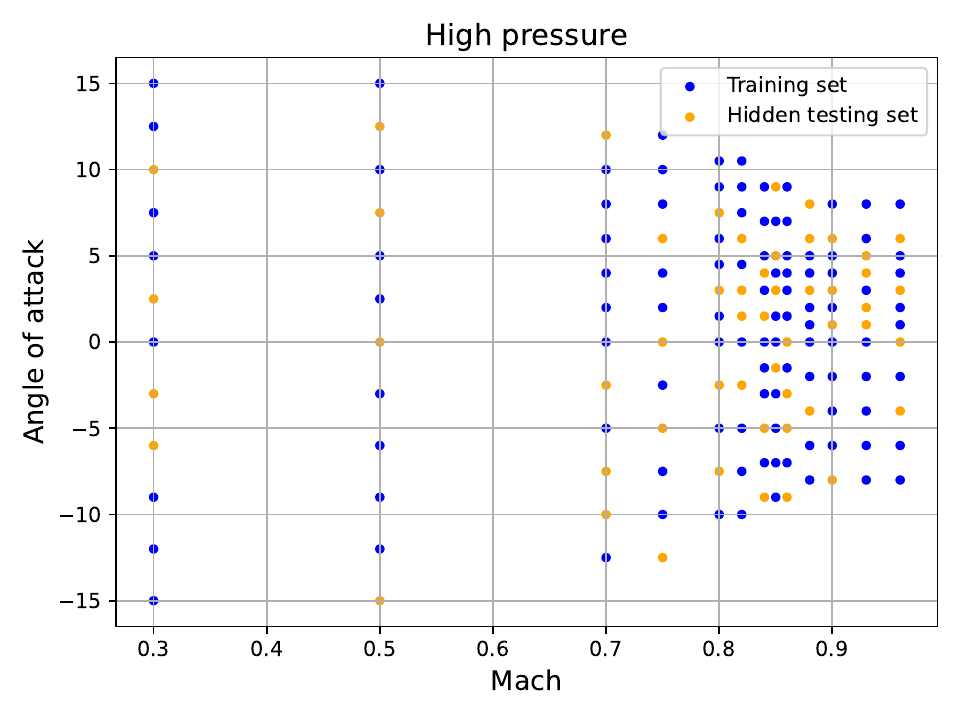}
        \label{fig:doe_high_pi}
    \end{subfigure}
    \caption{Training and hidden testing set repartition among the 468 different simulations of the database.}
    \label{fig:doe}
\end{figure}

The objective of the challenge is to predict the pressure and friction coefficients on the surface of the NASA CRM configuration under varying aerodynamic conditions. The dataset consists of 468 RANS simulations generated using the Spalart–Allmaras turbulence model. Among the 468 simulations, $n_{tr} = 312$ are made available for model development and training, while the remaining $n_{te}=146$ are retained as a hidden test set (see Figure \ref{fig:doe}). Predictions on these unseen configurations are evaluated exclusively through the Codabench platform, ensuring an unbiased assessment of model generalization.

Each simulation provides values at $n_p = 260, 774$ surface points, defined by their spatial coordinates and surface normals. The flow conditions are parameterized by the Mach number, Angle of Attack (AoA), and stagnation pressure $p_i$. In the CRM-WBPN database, the Reynolds number is uniquely determined from these variables and therefore does not constitute an additional degree of freedom in the parameter space~\cite[Section~2.2]{peter2025crm}. The dataset covers a broad spectrum of aerodynamic regimes, ranging from subsonic to near-stall conditions, and includes transonic cases characterized by complex and highly diverse flow behaviors.
The quality of the CFD solutions was assessed through confidence-factor metrics derived from the convergence history of both the residuals and the drag coefficient. Most simulations exhibit a satisfactory level of convergence; however, reduced confidence was observed for a subset of configurations corresponding to high Angles of Attack (AoAs) and low Mach numbers. To account for this variability, each simulation is associated with a confidence weight $w_f$ equal to 1.0 for well-converged cases and 0.5 for lower-confidence solutions. These weights are subsequently incorporated into the computation of the challenge evaluation metrics (Section~\ref{sec:metrics}), so reducing the influence of simulations affected by numerical convergence uncertainties.
An additional characteristic of the benchmark is that the computational mesh used to generate the CFD solutions is not provided. As a result, surrogate models must be designed independently of mesh connectivity, which strongly influences the choice of machine learning architectures and feature representations.

Formally, the objective is to predict the aerodynamic surface fields:
\begin{equation}
\mathbf{C} = (C_p, C_{fx}, C_{fy}, C_{fz})
\end{equation}
as a function of spatial location and flow parameters. Then, the model learns:
\begin{equation}
\mathbf{C}(x,y,z) = f_\theta(x,y,z,n_x,n_y,n_z,\mathrm{AoA},M_\infty,p_i)
\end{equation}
where $(x,y,z)$ are spatial coordinates, $(n_x,n_y,n_z)$ the surface normal, and $(\mathrm{AoA},M_\infty, p_i)$ the aerodynamic conditions.

\subsection{Evaluation Metrics} \label{sec:metrics}

Model performance is assessed using two complementary metrics: the coefficient of determination ($R^2$) and the worst-case relative Mean Absolute Error ($\mathrm{wrMAE}$). While the former measures the overall predictive accuracy of the surrogate model across the entire test set, the latter evaluates its robustness by focusing on the most challenging test configuration.
For a given aerodynamic quantity $y$ and its prediction $\hat{y}$, the coefficient of determination is defined as
\begin{equation*}
    R^2_y = 1 - \frac{\sum\limits_{b=0}^{n_{te}} \sum\limits_{i=0}^{n_p} w_f^b \left(y_{b,i} -\hat{y}_{b,i} \right)^2}{\sum\limits_{b=0}^{n_{te}} \sum\limits_{i=0}^{n_p} w_f^b \left(y_{b,i} -\bar{y} \right)^2},
    \label{eq:r2}
\end{equation*}
with
\begin{equation*}
    \bar{y} = \frac{1}{n_{te}n_p}\sum\limits_{b=0}^{n_{te}} \sum\limits_{i=0}^{n_p} y_{b,i},
\end{equation*}
where $n_{te}$ denotes the number of test simulations, $n_p$ the number of surface points, and $w_f^b$ the confidence weight associated with simulation $b$. The metric is computed independently for each target variable, yielding $R^2_{C_p}$, $R^2_{Cf_{x}}$, $R^2_{Cf_{y}}$, and $R^2_{Cf_{z}}$. The global coefficient of determination is then obtained by averaging the four quantities,
\begin{equation*}
    R^2 = \frac{ R^2_{C_p} + R^2_{Cf_{x}} + R^2_{Cf_{y}}+ R^2_{Cf_{z}}}{4}.
\end{equation*}
To complement this global measure of accuracy, the challenge introduces a worst-case relative Mean Absolute Error (wrMAE). For a given simulation $b$, the relative mean absolute error is then given by
\begin{equation*}
    \mathrm{rMAE}_y^b = \frac{\sum\limits_{i=0}^{n_p} |y_{i, b} - \hat{y}_{i, b}|}{\sum\limits_{i=0}^{n_p} |y_{i, b}|}.
    \label{eq:rmae}
\end{equation*}
The corresponding worst-case relative mean absolute error is then defined as
\begin{equation*}
    \mathrm{wrMAE}_y = \max_{b\in[1, n_{te}]} \mathrm{rMAE}_y^b.
    \label{eq:wrmae}
\end{equation*}
Similarly to the $R^2$ metric, $\mathrm{wrMAE}$ is computed independently for each predicted quantity, yielding $\mathrm{wrMAE}_{C_p}$, $\mathrm{wrMAE}_{Cf_x}$, $\mathrm{wrMAE}_{Cf_y}$, $\mathrm{wrMAE}_{Cf_z}$, and then averaged over the four aerodynamic variables,
\begin{equation*}
    \mathrm{wrMAE} = \frac{\mathrm{wrMAE}_{C_p} + \mathrm{wrMAE}_{Cf_{x}} + \mathrm{wrMAE}_{Cf_{y}} + \mathrm{wrMAE}_{Cf_{z}}}{4}.
\end{equation*}
The final challenge score combines both metrics according to
\begin{equation}
\text{Score} = 5 \times \mathrm{R}^2 + 5 \times (1 - \mathrm{wrMAE}).
\end{equation}
The maximum achievable score is 10, corresponding to perfect predictions with $R^2=1$ and $\mathrm{wrMAE}=0$.
\paragraph{Metrics interpretation.}
The evaluation protocol is designed to jointly assess predictive accuracy and robustness. The $R^2$ metric quantifies the fraction of variance explained by the surrogate model and therefore measures its overall ability to reproduce the aerodynamic fields over the complete test set. Values close to one indicate excellent agreement with the reference CFD solutions, whereas values near or below zero correspond to models that perform no better than a constant predictor.

In contrast, the $\mathrm{wrMAE}$ metric focuses exclusively on the most difficult test configuration by considering the largest relative error observed across all simulations. As a result, minimizing $\mathrm{wrMAE}$ requires the model to maintain satisfactory performance even in regions of the parameter space associated with complex flow phenomena, such as transonic shocks or near-stall conditions. Optimizing only for global accuracy may therefore be insufficient, since large errors on a small number of challenging configurations can significantly degrade the final score.

From a statistical perspective, the combined score emphasizes models that not only capture the central tendency and variance of the data distribution, as reflected by the $R^2$ metric, but also provide robust predictions in the tails of the distribution. The benchmark therefore rewards surrogate models capable of balancing average predictive performance with reliability on the most demanding aerodynamic conditions.

\subsection{Obstacles}
The challenge presents several difficulties that make the development of accurate surrogate models particularly demanding. These obstacles arise from both the physical complexity of the underlying aerodynamic problem and the practical constraints imposed by the competition framework.

A first challenge is the limited amount of training data available relative to the complexity of the target problem. Although each simulation contains aerodynamic quantities at more than $2.6 \times 10^5$ surface locations, only 312 CFD configurations are provided for model development. The dataset must therefore capture a high-dimensional mapping between aerodynamic operating conditions and wall distributions using a relatively small number of independent flow realizations. This situation is particularly challenging given the large variability of the flow physics represented in the database.

The diversity of aerodynamic regimes constitutes another major difficulty. The dataset covers Mach numbers ranging from 0.3 to 0.96 and AoA between $-15^\circ$ and $15^\circ$, spanning subsonic, transonic, and near-stall conditions. As a consequence, the surrogate model must simultaneously learn smooth flow regimes and highly nonlinear phenomena such as shock waves, strong pressure gradients, flow separation, and the onset of stall. Several high-angle-of-attack configurations also exhibit increased drag coefficient fluctuations, indicating emerging flow instabilities and reduced numerical convergence confidence. Accurately capturing such heterogeneous flow behaviors within a single model is inherently challenging.

The geometric complexity of the NASA/Boeing Common Research Model further increases the difficulty of the task. Unlike simplified wing-only benchmarks, the CRM-WBPN configuration includes the wing, fuselage, pylon, and nacelle components, resulting in a complex three-dimensional geometry with intricate local flow interactions. Moreover, the aircraft surface cannot be naturally mapped onto a regular Cartesian grid without introducing significant distortions \cite{massegur2024graph}. This prevents the direct application of conventional CNN architectures that have demonstrated strong performance on structured image-like data.

An additional constraint is the absence of the CFD mesh from the released dataset. Since mesh connectivity information is unavailable, machine-learning approaches based on graph or mesh operators cannot be directly applied without reconstructing an approximate connectivity structure. Surrogate models must therefore rely exclusively on geometric descriptors, surface coordinates, and flow parameters, which naturally favors mesh-independent representations.

The characteristics of the target fields themselves also contribute to the complexity of the problem. Pressure and skin-friction distributions exhibit spatial structures spanning multiple scales, ranging from large smooth regions to highly localized discontinuities associated with shocks and separation phenomena. Such multiscale behavior is known to be difficult for standard neural networks due to their tendency to preferentially learn low-frequency features, a phenomenon commonly referred to as spectral bias.

Finally, the competition framework imposes additional methodological constraints. Participants were limited to twenty submissions on the hidden test set throughout the duration of the challenge. Consequently, model selection and hyperparameter optimization could not rely on repeated leaderboard probing. A possible solution is the construction of a sufficiently representative local validation strategy but this further reduced the amount of data available for training, introducing a trade-off between model development and reliable performance estimation. Furthermore, the evaluation score combines a global accuracy metric ($R^2$) with a worst-case error criterion ($\mathrm{wrMAE}$), requiring models not only to achieve strong average performance but also to remain robust on the most challenging flow configurations. This dual objective significantly increases the difficulty of the challenge and strongly influences the design of both training objectives and model architectures.
\section{Reference Baseline Models} \label{sec:baselines}
To establish reference performance levels for the CRM-WBPN benchmark, Peter et al.~\cite{peter2025crm} evaluated several machine-learning approaches, including both pointwise and global regression strategies. Among the organizer-provided baselines, the best-performing model was a global Multi-Layer Perceptron (Global MLP), also denoted as a \textit{Modewise MultiLayer Perceptron} in the original publication~\cite[Section~5.1]{peter2025crm}. A pointwise regression model was also investigated and is arguably the baseline most closely related to the neural-field formulation proposed in this work. Nevertheless, its performance remained significantly below that of the Global MLP.

The benchmark study considered a total of seven surrogate modeling approaches. A subset of these methods had previously been developed and evaluated by the organizers in~\cite{peter2024} on a closely related aerodynamic surrogate-modeling benchmark involving a wing geometry subjected to varying operating conditions. The pointwise methods included a standard MLP, a $\lambda$-DNN architecture specifically designed to better couple geometric information with aerodynamic operating conditions, and a decision-tree regressor. The global approaches comprised the Global MLP, a $k$-Nearest-Neighbor (kNN) interpolation method operating in the input parameter space, an IsoMap-based dimensionality reduction combined with Radial Basis Function (RBF) interpolation, and a reduced-order modeling strategy based on Proper Orthogonal Decomposition (POD) coupled with RBF regression. Together, these baselines cover a broad range of surrogate modeling philosophies, from local pointwise prediction to global field reconstruction through nonlinear manifold learning and reduced-order representations.
\subsection{Global MLP Baseline}
The Global MLP adopts a direct field-regression strategy. Instead of predicting aerodynamic quantities independently at each surface location, the model learns a mapping from the aerodynamic operating conditions to the complete wall distribution of a given target quantity. For each aerodynamic field, a dedicated neural network is trained to approximate the mapping
\begin{equation}
f_\theta :(M_\infty,AoA,p_i) \rightarrow \mathbf{C},
\end{equation}
where $\mathbf{C}$ denotes the full field containing the values of either $C_p$, $C_{f_x}$, $C_{f_y}$, or $C_{f_z}$ at all surface locations. Consequently, four independent neural networks are required to predict the complete set of challenge outputs.
Unlike pointwise regressors, which learn local relationships at individual surface locations, the Global MLP directly models the entire aerodynamic field as a high-dimensional output vector. This formulation allows the network to exploit correlations between distant regions of the aircraft surface and implicitly learn global flow structures. \\
Because the output dimension exceeds $2.6\times10^5$ values per simulation, direct field prediction leads to an exceptionally large regression problem. Following a dedicated hyperparameter search, Peter et al.~\cite{peter2025crm} selected a Global MLP architecture with hidden-layer dimensions $(75, 120, 1226, 16490)$. Owing to the very large output layer, each network contains approximately $4\times10^9$ trainable parameters. As four independent networks are required to predict $C_p$, $C_{f_x}$, $C_{f_y}$, and $C_{f_z}$, the overall baseline comprises more than $1.7\times10^{10}$ parameters, making it several orders of magnitude larger than the neural-field models considered in this work.
\subsection{Strengths and Limitations}
The Global MLP baseline presents several attractive characteristics. First, it directly predicts the target fields without requiring any intermediate representation, dimensionality reduction technique, or geometric preprocessing. The approach therefore remains conceptually simple and straightforward to implement.

Second, by treating the complete aerodynamic field as a single output object, the model can implicitly learn large-scale correlations across the aircraft surface. Such correlations are difficult to capture with purely local regression strategies.

However, this formulation also exhibits several limitations. Most importantly, the geometry itself is not explicitly provided to the network. The model only receives the aerodynamic operating conditions as inputs and must therefore infer the entire wall distribution from a low-dimensional parameter space. Consequently, all geometric information is embedded implicitly within the training data and the learned network weights.

Furthermore, the direct prediction of more than $260,000$ output values leads to a very large regression problem. While the network can successfully capture dominant flow trends, representing localized phenomena such as shock waves, separation regions, and sharp pressure gradients remains challenging. These features often correspond to highly nonlinear and localized variations of the aerodynamic fields, which are difficult to reconstruct accurately through a purely global parameter-to-field mapping.
\subsection{Performance Analysis}
Despite its conceptual simplicity, the Global MLP constitutes a remarkably strong baseline. According to the results reported by Peter et al.~\cite{peter2025crm}, it achieved the best performance among the seven baseline models evaluated by the organizers, both in terms of the global accuracy metric ($R^2$) and the worst-case relative error. The authors attribute this performance, at least partially, to the exceptionally large number of trainable parameters of the model, which may provide sufficient capacity to learn the highly nonlinear relationship between operating conditions and aerodynamic wall distributions.

Nevertheless, the qualitative analysis presented by Peter et al.~\cite{peter2025crm} reveals several limitations of the Global MLP. Although the global error metrics are relatively high, some important flow features remain poorly reconstructed. In particular, the pressure coefficient and skin-friction predictions fail to accurately reproduce the transonic shock observed on the upper wing surface for certain operating conditions. These observations suggest that strong global metrics do not necessarily guarantee an accurate representation of localized aerodynamic phenomena, especially when sharp discontinuities and spatial high-frequency structures are involved.
The methodology proposed in this paper, introduced in Section~\ref{sec:methodology}, is explicitly designed to address these limitations by improving the reconstruction of small-scale and localized flow structures.

\section{Methodology} \label{sec:methodology}
\begin{figure}[!ht]
    \centering
    \includegraphics[width=0.9\linewidth]{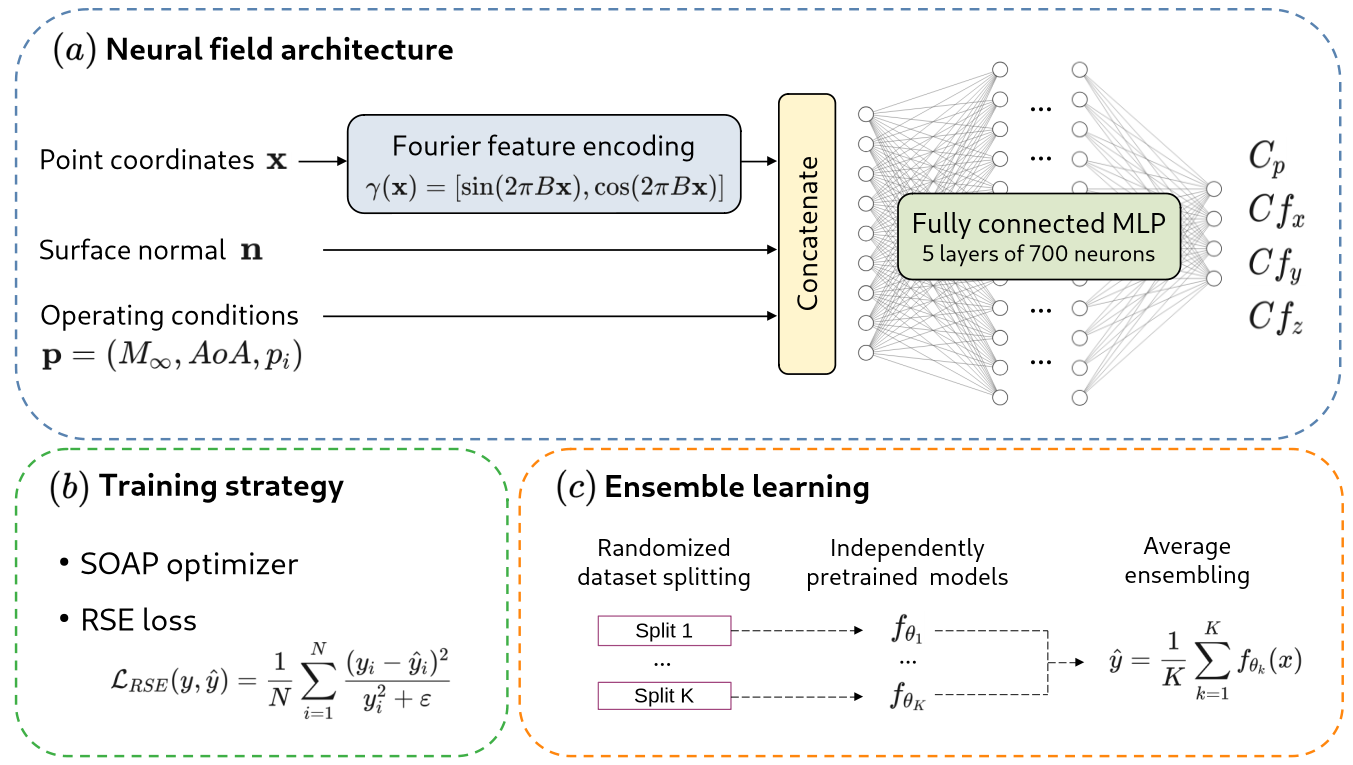}
    \caption{Methodology overview.}
    \label{fig:arch}
\end{figure}

%
The proposed final solution is based on a coordinate-based neural field representation trained to predict aerodynamic wall quantities directly from geometric coordinates and operating conditions. The approach was specifically designed to address the constraints of the CRM-WBPN benchmark, namely the absence of mesh connectivity information, the large number of surface points, and the presence of multiscale flow phenomena ranging from smooth pressure distributions to localized shock structures.
The methodology developed in this work is summarized in Figure \ref{fig:arch}.
\subsection{Neural Field Formulation} \label{sec:neural_field}
Neural fields, also referred to as Implicit Neural Representations (INRs), model a physical quantity as a continuous function parameterized by a neural network. Instead of storing the values of a field on a predefined mesh or grid, the network directly learns a mapping from spatial coordinates to the corresponding field values \cite{essakine2024we}. This representation has recently emerged as a powerful paradigm for modeling complex signals and geometric objects, due to its ability to represent continuous functions with arbitrary spatial resolution.

The neural field formulation can naturally be extended by conditioning the network on additional parameters describing the physical system. In the context of surrogate modeling, these conditioning variables typically correspond to boundary conditions, material properties, or operating parameters \cite{xie2022neural}. Such coordinate-based representations have been successfully applied to the approximation of PDE solutions and operator learning problems on complex geometries \cite{serrano2023operator}. They also constitute the foundation of several Physics-Informed Neural Network (PINN) formulations, where the network output is constrained through the governing equations of the underlying physical system \cite{cuomo2022scientific}.

In the present work, let $\mathbf{x}=(x,y,z)$ denote the spatial coordinates of a point located on the aircraft surface, $\mathbf{n}=(n_x,n_y,n_z)$ the corresponding surface normal vector, and $\mathbf{p}=(M_\infty,AoA,p_i)$ the aerodynamic operating conditions. The objective is to learn the continuous mapping
\begin{equation}
    f_\theta : (\mathbf{x},\mathbf{n},\mathbf{p}) \longrightarrow \left(C_p(\mathbf{x}),C_{f_x}(\mathbf{x}),C_{f_y}(\mathbf{x}),C_{f_z}(\mathbf{x})\right),
\end{equation}
where $\theta$ denotes the trainable parameters of the neural network. The model therefore predicts the pressure coefficient and the three components of the skin-friction coefficient at any location on the aircraft surface.

\paragraph{Motivation.}
Several characteristics of the challenge data make neural fields particularly well suited to the problem. First, the dataset is provided as an unstructured collection of surface points associated with geometric descriptors, while the underlying CFD mesh connectivity is not available. As a result, methods that explicitly rely on mesh topology or graph connectivity cannot be directly applied. In contrast, neural fields operate directly on point coordinates and geometric features, without requiring any connectivity information. This mesh-independent formulation enables the modeling of physical fields on irregular and geometrically complex configurations, such as the CRM-WBPN geometry considered in this challenge, while remaining agnostic to the discretization. 

A second advantage of neural fields lies in their continuous representation of the solution. Although not required by the challenge evaluation protocol, the learned field can in principle be queried at arbitrary spatial locations, independently of the point distribution used during training. More importantly, the coordinate-based formulation provides a flexible framework for representing aerodynamic quantities exhibiting a wide range of spatial scales, from smooth variations over large portions of the aircraft surface to localized flow structures associated with shocks, strong pressure gradients, or separated regions.

Finally, conditioning the neural field on the aerodynamic operating conditions naturally leads to a parametric surrogate model of the flow solution manifold. From this perspective, the approach can be viewed as an instance of operator learning, where the network learns a continuous mapping from operating conditions to aerodynamic fields~\cite{kovachki2023neural, azizzadenesheli2024neural}. The resulting model is able to interpolate continuously within the parameter space and predict pressure and friction distributions for previously unseen flight conditions, which is precisely the objective of the challenge.
\subsection{Fourier Feature Encoding} \label{sec:fourier-features}
A well-known limitation of MLPs is their spectral bias \cite{rahaman2019spectral}, which causes them to preferentially learn low-frequency functions while struggling to represent high-frequency features. This limitation is particularly problematic for aerodynamic applications where shock waves and separation regions generate localized discontinuities and sharp gradients.
To alleviate this issue, the spatial coordinates are first transformed through a Fourier Feature Encoding (FFE) inspired by the positional encodings commonly used in neural fields. For an input coordinate vector $\mathbf{x}$, the encoded representation is given by
\begin{equation*}
    \gamma(\mathbf{x}) = \left[\sin(2\pi B\mathbf{x}), \cos(2\pi B\mathbf{x})\right],
\end{equation*}
where $B\in\mathbb{R}^{N_e \times 3}$ is a matrix of random frequencies whose entries are independently sampled from a zero-mean Gaussian distribution with variance $\sigma^2$. The parameter $\sigma^2$ controls the frequency spectrum represented by the FFE and therefore determines the range of spatial scales that can be captured by the model.
The encoded coordinates are subsequently concatenated with the aerodynamic operating conditions and normal before being processed by the neural network. This representation enriches the input space with multiple spatial frequencies and significantly improves the ability of the model to capture localized flow structures.
The encoding hyperparameters were selected empirically based on validation experiments. In particular, the variance of the frequency distribution was fixed to $\sigma^2 = 1$, while the encoding dimension was set to $N_e = 128$.
\subsection{Network Architecture}
The surrogate model is implemented as a fully connected MLP that receives as input the Fourier-encoded spatial coordinates (Section~\ref{sec:fourier-features}), the surface normal vector, and the aerodynamic operating conditions. The network jointly predicts the four target quantities $(C_p, Cf_{x}, Cf_{y}, Cf_{z})$. Adopting a multi-output formulation enables the model to exploit the physical correlations that exist between pressure and skin-friction fields, while avoiding the additional computational cost associated with training separate networks for each quantity.

\paragraph{Normalization.}
All input and output variables are normalized prior to training. The spatial coordinates are linearly scaled to the interval $[-1,1]$, while the surface normal vector is already normalized by construction. The Mach number $M_\infty$ and stagnation pressure $p_i$ are scaled to $[0,1]$, whereas the AoA is mapped to $[-1,1]$. Regarding the target variables, the pressure coefficient field is standardized to zero mean and unit variance, while the three skin-friction components are linearly scaled to the interval $[-1,1]$.
These normalization strategies were selected empirically based on validation experiments and were found to improve both training stability and predictive accuracy. In particular, bringing all variables to comparable numerical ranges facilitates optimization and reduces the risk of gradient imbalance between the different inputs and outputs.

\paragraph{MLP Architecture.}
The final architecture consists of a deep feed-forward network with five hidden layers, each containing 700 neurons, resulting in approximately $2.15\times10^6$ trainable parameters. A LeakyReLU activation function is applied after each hidden layer to introduce nonlinearity while maintaining stable gradient propagation throughout the network.
Several architectures with varying depths and widths were evaluated during the development process. While increasing the network size generally led to modest improvements in validation performance, the gains became marginal once the number of trainable parameters exceeded approximately $2.15\times10^6$. The selected architecture therefore represents a favorable compromise between model capacity, computational efficiency, and generalization performance.
\subsection{Loss Function} \label{sec:lossrse}
Preliminary experiments revealed that optimizing a conventional mean squared error objective produced models with strong $R^2$ scores but suboptimal performance with respect to the challenge ranking metric. Since the leaderboard score explicitly depends on the relative prediction error through the wrMAE metric, a better alignment between training and evaluation objectives was required.
The final model was therefore trained using a Relative Squared Error (RSE) loss defined as
\begin{equation*}
    \mathcal{L}_{RSE}(y, \hat{y}) = \frac{1}{N}\sum_{i=1}^{N} \frac{(y_i-\hat{y}_i)^2}{y_i^2 + \varepsilon},
\end{equation*}
where $\varepsilon$ is a small numerical constant introduced for stability, $y$ and $\hat{y}$ are the ground truth and prediction respectively.
Compared with the standard MSE, this formulation assigns a larger weight to relative prediction accuracy and was found to substantially improve the wrMAE metric, which ultimately resulted in higher leaderboard scores.
\subsection{Optimization Strategy}
Model training was performed using the SOAP optimizer~\cite{vyas2025soap}, combined with a learning-rate schedule consisting of an initial linear warm-up phase followed by cosine annealing~\cite{loshchilov2017sgdr}. All experiments were conducted on a single NVIDIA A100 GPU. The validation loss was monitored throughout training, and early stopping was employed to prevent overfitting and reduce unnecessary computation once the validation performance ceased to improve. The strategy used to partition the available data into training and validation subsets is described in Section~\ref{sec:kfold}.
The initial learning rate was set to $10^{-2}$ and the batch size to $2\times n_p$, where $n_p$ denotes the number of surface points per configuration. These hyperparameters were selected empirically based on validation performance. The early-stopping patience was fixed to 20 epochs, meaning that training was terminated after 20 consecutive epochs without improvement in the validation loss.

The SOAP optimizer was retained following comparative experiments against AdamW. While both optimizers achieved similar levels of predictive accuracy, SOAP consistently provided slightly improved validation performance with only a modest increase in computational cost. Unlike AdamW, which relies exclusively on first-order gradient information, SOAP incorporates a second-order preconditioning strategy based on an approximation of the Hessian matrix. This additional curvature information improves the conditioning of the optimization problem and can lead to faster convergence and enhanced optimization accuracy.
\subsection{Ensemble Learning} \label{sec:ensemble}
Although the single neural-field model achieved strong predictive performance, a noticeable discrepancy remained between validation and hidden-test results, particularly for the worst-case error metric. 
To improve robustness, multiple independently trained neural fields can be combined through ensemble averaging (see, e.g., \cite{naftaly1997optimal}). Let $f_{\theta_1},\ldots,f_{\theta_K}$ denote $K$ independently trained models. The final prediction is then obtained as
\begin{equation}
    \hat{y} = \frac{1}{K} \sum_{k=1}^{K} f_{\theta_k}(x).
\end{equation}
Averaging predictions reduces variance and mitigates individual model errors, leading to improved generalization and lower worst-case prediction errors.
\subsection{K-Fold Cross-Validation Ensemble} \label{sec:kfold}
The final competition submission relies on a cross-validation ensemble strategy. The 312 available simulations are partitioned into five folds. For each fold, a dedicated training-validation split is constructed and several neural fields are trained using different random initializations.

The 20 best-performing models across all folds are retained and combined into a final ensemble, corresponding to approximately $7.3 \times 10^7$ trainable parameters.

This strategy increases the diversity of the ensemble while allowing every simulation to contribute to model training. The resulting predictor achieved the highest score on the final hidden test set and constitutes the final solution presented in this work.

\section{Progressive Development and Ablation Study} \label{sec:ablation}
This section analyzes the successive methodological choices that led to the final winning solution. Starting from a simple neural-field baseline, each modification was introduced to address a specific limitation identified through validation experiments. The objective is not only to justify the final architecture but also to provide practical insights into the design of surrogate models for aerodynamic wall-field prediction.
\subsection{Initial Neural Field Prototype}
The development process began with a simple coordinate-based neural field implemented as a MLP, following the reasoning of the formulation introduced in Section~\ref{sec:neural_field}. Adopting a simple MLP architecture allowed us to establish a lightweight and easily reproducible baseline, facilitating rapid experimentation and the progressive evaluation of subsequent architectural improvements. The initial prototype consisted of a fully connected MLP with five hidden layers of 128 neurons each and LeakyReLU activation functions, resulting in approximately $6.7\times10^4$ trainable parameters. A single network was trained to jointly predict the four target quantities, as preliminary experiments outperformed the alternative strategy of training one model per field. This behavior suggests that jointly learning the pressure and skin-friction fields enables the network to exploit the physical correlations between these quantities. The model was trained using a weighted MSE loss, with the relative weights between the four outputs selected empirically, and optimized using AdamW. A total of 260 simulations were used for training, while the remaining 52 configurations were reserved for validation.

The performance of this initial prototype is summarized in Table~\ref{tab:ablation_summary}. Although the model produced encouraging predictions ($R^2 > 0.90$) despite its modest size, it remained substantially less accurate than both the organizer's point-wise MLP and Global MLP baselines, which comprise approximately $1.25\times10^5$ and $1.7\times10^{10}$ trainable parameters, respectively. This performance gap initially suggested that increasing the model capacity could improve predictive accuracy.
More importantly, qualitative inspection of the predicted fields revealed a limited ability to reproduce localized flow structures, particularly shock waves. This observation highlighted the spectral limitations of conventional MLPs and motivated the introduction of Fourier feature encoding, described in the next subsection.
\subsection{Fourier Feature Encoding}
The initial neural-field prototype demonstrated that a simple coordinate-based MLP was capable of learning the overall structure of the aerodynamic fields. However, visual inspection of the predictions consistently revealed difficulties in reproducing localized flow structures. This limitation was observed despite increasing the network size, suggesting that the bottleneck was not solely related to model capacity but also to the spectral bias mentioned in Section \ref{sec:fourier-features}. 
Since aerodynamic wall quantities simultaneously exhibit smooth global trends and localized discontinuities, particularly in transonic regimes, enriching the spatial representation appeared more promising than only increasing the number of trainable parameters.

To address the spectral bias, the Cartesian coordinates were replaced by the Fourier Feature Encoding (FFE) introduced in Section~\ref{sec:fourier-features}. The same encoding was also applied to the surface normal vectors, but as it yielded no measurable improvement in predictive performance, the original representation was retained.

The introduction of FFE substantially improved both the $R^2$ score and the $\mathrm{wRMAE}$, as summarized in Table~\ref{tab:ablation_summary}, resulting in an increase of the overall score from $7.80$ to $8.34$. The most pronounced qualitative improvements were observed in regions exhibiting localized flow phenomena, including transonic shock waves and abrupt variations in the skin-friction fields, where sharp discontinuities and gradients were reproduced much more accurately. In addition to the ability of FFE to alleviate MLP spectral bias, FFE considerably increases the expressive power of the model while preserving the network architecture and adding only a negligible number of trainable parameters.

\subsection{Increasing Model Capacity}

Following the introduction of FFE, the next objective was to determine whether the model remained capacity-limited. The network width was therefore progressively increased while keeping the overall architecture unchanged. As summarized in Table~\ref{tab:ablation_summary}, increasing the model capacity consistently improved predictive performance, raising the overall score from $8.34$ to $8.45$ and yielding improvements in both the $R^2$ and $\mathrm{wRMAE}$ metrics for all predicted quantities except $\mathrm{wrMAE}_{Cf_y}$. 

While larger networks were better able to capture the highly nonlinear mapping between the geometric coordinates, operating conditions, and aerodynamic fields (final single-model architecture containing approximately $2.15\times10^6$ trainable parameters), further increasing the model size resulted in marginal performance gains but substantial increase in the computational cost. This architecture was therefore retained as the best compromise between predictive accuracy and computational efficiency.
\subsection{Metric-Aligned Loss Function}
Despite the improvements obtained through FFE and increased model capacity, the validation analysis revealed an important discrepancy between the optimization objective and the competition evaluation metric. Models trained with a conventional MSE loss achieved excellent $R^2$ values but remained comparatively weak on the wrMAE metric. This behavior can be explained by the fact that the challenge score explicitly incorporates a relative error measure. Large errors on a small number of challenging configurations therefore have a disproportionate impact on the final ranking, even when the global prediction accuracy remains high.

To better align the training objective with the evaluation protocol, the mean squared error was replaced by a Relative Squared Error (RSE) loss as described in Section \ref{sec:lossrse}. This formulation normalizes the prediction error by the magnitude of the target quantity, encouraging the network to minimize relative rather than absolute errors. It also places the loss contributions of the different target fields on a comparable scale, eliminating the need for manually tuned weighting coefficients.

The impact of this modification was immediate. While the $R^2$ metric remained essentially unchanged (from $0.965$ to $0.967$), a substantial reduction of the wrMAE was observed decreasing from $0.276$ to $0.251$. The wrMAE was improved for all the fields except for $Cf_x$ which remained unchanged. The overall validation score increased from approximately $8.45$ to $8.58$, demonstrating that loss-function design can be as important as architectural improvements when the evaluation metric emphasizes robustness.
\subsection{Ensemble Learning} \label{sec:progr:ensemble}

Although the single neural-field model achieved strong validation performance, a persistent gap remained between the validation and hidden-test scores. This discrepancy was primarily attributable to the $\mathrm{wrMAE}$ metric, indicating that a small number of hidden-test predictions were less accurate. 
To improve the robustness of the predictions, an ensemble learning strategy was therefore introduced, as described in Section~\ref{sec:ensemble}. Multiple neural-field models sharing the same architecture and training dataset, but initialized with different random seeds, were trained independently. The final prediction was obtained by averaging the outputs of the individual models.
The underlying hypothesis is that the prediction errors of independently initialized models are only partially correlated. Consequently, configurations that are poorly predicted by one model may be better approximated by another, allowing the averaging process to reduce the influence of individual prediction failures. From a statistical perspective, ensemble averaging reduces the variance of the estimator while preserving its bias, resulting in more robust predictions and improved generalization on previously unseen operating conditions.

A seven-model ensemble, corresponding to approximately $1.5\times10^7$ trainable parameters, was initially constructed. As reported in Table~\ref{tab:ablation_summary}, ensemble averaging improved both the $R^2$ and $\mathrm{wrMAE}$ metrics, increasing the hidden-test score from $8.58$ to $8.65$. Nevertheless, the improvement remained moderate, suggesting that the dominant source of error was no longer the random initialization of the models but rather the limited diversity of the training data. Since all ensemble members were trained on the same train-validation split, they were exposed to the same distribution of aerodynamic configurations and therefore tended to share similar weaknesses. This observation motivated the investigation of a $k$-fold cross-validation strategy, presented in Section~\ref{sec:progr:kfold}, to increase the diversity of the ensemble while exploiting the entire training dataset.

\subsection{K-Fold Cross-Validation Ensemble} \label{sec:progr:kfold}
A detailed analysis of the hidden-test predictions revealed that the largest errors were concentrated in a small number of configurations characterized by relatively high Mach numbers and large Angles of Attack ($M_\infty = 0.85$, $AoA = 8.0^\circ$, $p_i = 4.0\times10^5$ Pa for example). These regions correspond to complex transonic flow conditions where shocks and partial flow separation become increasingly prominent.
This observation suggested that part of the remaining error could be attributed to limitations of the training-validation split. Given the relatively small number of available simulations, certain regions of the parameter space may be underrepresented in a single training partition.

To mitigate this effect, a five-fold cross-validation strategy was adopted. The 312 available simulations were partitioned into five different train-validation configurations. For each fold, 8 neural-field models were trained using different random initializations. The best-performing models from all folds were subsequently aggregated into a single ensemble.
This approach offers two complementary advantages. First, every simulation contributes to model training in multiple folds, resulting in a more efficient use of the available data. Second, the diversity of the resulting ensemble is significantly increased because it contains models that are exposed to a different subset of training examples.

The final ensemble combines twenty neural-field models corresponding to approximately $73$ million trainable parameters. This strategy produced a significant improvement, reducing the wrMAE from 0.241 to 0.210 while simultaneously increasing the $R^2$ score. The resulting leaderboard score reached 8.8141 and ultimately secured first place in the challenge.
\subsection{Summary of Performance Improvements}
Table~\ref{tab:ablation_summary} summarizes the successive improvements obtained throughout the development process. The results highlight that the winning solution emerged from the combination of several complementary ingredients: improved spatial encoding through FFE, tailored model capacity, metric-aligned optimization, ensemble regularization, and cross-validation-based data utilization.

These results suggest that, for the CRM-WBPN benchmark, improvements in robustness and generalization were ultimately more beneficial than further increases in model complexity. In particular, the most significant gains were obtained through techniques specifically targeting the worst-case error metric, which plays a central role in the challenge evaluation protocol.

\begin{table}[ht]
    \centering
    \caption{Progressive improvement of the solution throughout the development process.}
    \label{tab:ablation_summary}
    \begin{tabular}{l|c|c|c|c|c|c|}
        \cline{2-7}
         & 
            \rotatebox[origin=l]{90}{Initial neural field prototype} &     
            \rotatebox[origin=l]{90}{ + Fourier Features Encoding (FFE)} &                
            \rotatebox[origin=l]{90}{ + Increased model capacity} &        
            \rotatebox[origin=l]{90}{ + Relative squared error loss} &     
            \rotatebox[origin=l]{90}{ + Ensemble learning} &               
            \rotatebox[origin=l]{90}{ + K-fold cross-validation ensemble } 
            \\
        \hline
        Score $\uparrow$ & 7.80 & 8.34 & 8.45 & 8.58 & 8.65 & \textbf{8.81} \\
        \hline
        $R^2$  $\uparrow$ & 0.908 & 0.943 & 0.965 & 0.967 & 0.970 & \textbf{0.973} \\
        \hline
        $R^2_{C_p}$ & 0.964 & 0.978 & 0.979 & 0.980 & 0.982 & 0.983 \\
        $R^2_{Cf_{x}}$ & 0.891 & 0.922 & 0.956 & 0.958 & 0.962 & 0.966 \\
        $R^2_{Cf_{y}}$ & 0.890 & 0.931 & 0.960 & 0.963 & 0.966 & 0.969 \\
        $R^2_{Cf_{z}}$ & 0.888 & 0.942 & 0.966 & 0.968 & 0.971 & 0.973 \\
        \hline
        $\mathrm{wrMAE}$  $\downarrow$ & 0.349 & 0.276 & 0.276 & 0.251 & 0.241 & \textbf{0.210} \\
        \hline
        $\mathrm{wrMAE}_{C_p}$ & 0.286 & 0.200 & 0.171 & 0.162 & 0.160 & 0.157 \\
        $\mathrm{wrMAE}_{Cf_{x}}$ & 0.236 & 0.194 & 0.179 & 0.179 & 0.173 & 0.170 \\
        $\mathrm{wrMAE}_{Cf_{y}}$ & 0.436 & 0.393 & 0.471 & 0.409 & 0.398 & 0.315 \\
        $\mathrm{wrMAE}_{Cf_{z}}$ & 0.439 & 0.317 & 0.282 & 0.253 & 0.233 & 0.197 \\
        \hline
    \end{tabular}
\end{table}
\subsection{Alternative Approaches Investigated}
Several alternative modeling strategies were explored before arriving at the final approach presented in this paper. Although some of these approaches appeared promising from a theoretical perspective, none ultimately provided a favorable trade-off between predictive performance, implementation complexity, and robustness compared with the final neural-field ensemble. This section summarizes the main directions investigated and the lessons learned from these exploratory experiments.  

\paragraph{Mesh reconstruction and graph neural networks.}

Graph Neural Networks (GNNs) constitute a natural candidate for CFD surrogate modeling, as they can exploit mesh connectivity information to propagate geometric and physical information across neighboring surface elements. Several recent studies have demonstrated the effectiveness of graph-based operators for learning aerodynamic quantities directly on unstructured meshes \cite{belbute2020combining, li2024physics}.

However, the CRM-WBPN benchmark \cite{peter2025crm} does not provide the computational mesh used to generate the CFD solutions. Only point coordinates and surface normals are available. Applying graph-based methods therefore requires reconstructing an approximate connectivity structure from the point cloud. Preliminary investigations were conducted using nearest-neighbor graphs and geometric proximity criteria. While these approaches produced visually plausible connectivity patterns, the resulting graphs remained only rough approximations of the original CFD mesh and introduced additional hyperparameters related to graph construction.

Given the limited number of available simulations and the substantial implementation effort required, graph-based approaches were not pursued further. In contrast, neural fields operate directly on point coordinates and naturally avoid the need for mesh reconstruction.

\paragraph{Parametric zonal decomposition strategies.}
A natural hypothesis was that the mapping between operating conditions and aerodynamic fields could be simplified by partitioning the parameter space into regions associated with different flow regimes and training a dedicated surrogate model for each region. Such an approach could reduce the complexity of the learning task by allowing each model to specialize in a narrower range of aerodynamic behaviors. Several partitioning strategies based on the AoA were therefore investigated. A first decomposition separated negative and positive incidence configurations ($AoA < -1^\circ$ and $AoA \geq -1^\circ$). A second strategy exploited prior knowledge of the CFD database by distinguishing three regions: $AoA < -4^\circ$, $-4^\circ \leq AoA \leq 5^\circ$, and $AoA > 5^\circ$. These thresholds were selected after analyzing the standard deviation of the drag coefficient during the last 20\% of the CFD iterations, which remains low within the central range but increases markedly at larger positive and negative incidences, indicating the progressive onset of flow instabilities and partial stall conditions.

Although this decomposition reduced the variability of the flow within each subdomain, it also substantially decreased the number of training simulations available for each individual model. Given the already limited size of the dataset, the loss of training data outweighed the potential benefit of specialization, and no consistent improvement was observed on the validation score. Consequently, parametric zonal decomposition strategies were not retained in the final methodology.

\paragraph{Alternative coordinate encodings.}
The representation of spatial coordinates was identified early as a critical component of the surrogate model. Several coordinate encoding strategies were considered, including direct coordinate inputs and learnable feature transformations.
Direct coordinate regression using a standard MLP consistently underperformed compared with FFE. Learnable frequency encoding \cite{li2021learnable} were also considered as a potential extension of the FFE approach. While theoretically appealing, preliminary experiments did not reveal a clear advantage over fixed FFE. Given their simplicity, computational efficiency, and strong empirical performance, fixed FFE were ultimately retained.



\paragraph{Design insights from exploratory investigations.}
Several conclusions emerge from these exploratory investigations. First, the absence of mesh connectivity strongly favors mesh-independent representations and limits the practical applicability of graph-based methods. Second, the ability to capture high-frequency spatial features is essential for accurately reproducing transonic flow phenomena, making coordinate encoding a critical design choice. Third, the challenge evaluation metric places a strong emphasis on robustness, which makes metric-aligned loss functions and ensemble strategies particularly effective. Finally, the results suggest that, for this benchmark, careful optimization of relatively simple neural-field models is more beneficial than introducing increasingly sophisticated architectures.

These observations provide useful guidance for future work on the CRM-WBPN dataset and, more broadly, for the design of machine-learning surrogates operating on complex aerodynamic geometries under limited-data conditions.

\section{Final Results} \label{section:results}
This section evaluates the performance of the proposed methodology (see Section \ref{sec:methodology}) on the hidden test set of the ONERA CRM Wall Distribution Regression Challenge and compares its results with those of the organizer-provided baseline, which ranked second on the leaderboard.

Table~\ref{tab:final_results} compares the performance of the proposed approach with the best organizer-provided baseline introduced by Peter et al.~\cite{peter2025crm} and the more recent FluidFlow model, independently evaluated on the same ONERA CRM-WBPN benchmark by Ramos et al.~\cite{ramos2026fluidflow}.

\paragraph{Performance comparison.}
The results demonstrate that the proposed methodology consistently outperforms the Global MLP baseline across all evaluation metrics. Improvements are observed not only for the average coefficient of determination but also for the worst-case relative error, indicating that the model achieves a better compromise between global predictive accuracy and robustness on the most challenging aerodynamic configurations. This behavior is particularly important since the competition score explicitly rewards methods capable of maintaining high accuracy on difficult operating conditions.

\paragraph{Computational cost.}
Despite relying on an ensemble of neural fields, the proposed approach remains computationally tractable. Each individual model contains approximately $2.15\times10^6$ trainable parameters, resulting in roughly $7.3\times10^7$ parameters for the final model ensemble. This remains several orders of magnitude smaller than the Global MLP baseline, which contains approximately $1.7\times10^{10}$ trainable parameters while achieving lower predictive performance. The proposed methodology therefore provides a parameter-efficient surrogate model for aerodynamic field prediction.

\begin{table*}[t]
    \centering
    \caption{Performance comparison between the proposed methodology and the best organizer-provided baseline on the hidden test set. FluidFlow results are taken from \cite{ramos2026fluidflow}, where the model was independently evaluated using the official ONERA CRM-WBPN train/test split. Empty cells indicate metrics not reported in the original publication.}
    \label{tab:final_results}
    \begin{tabular}{l|c|c|c|}
        \cline{2-4}
         & 
            \textbf{Proposed methodology} &
            Global MLP (Peter et al.~\cite{peter2025crm}) &
            FluidFlow (DiT) \cite{ramos2026fluidflow}
            \\
        \hline
        Score $\uparrow$ & \textbf{8.81} & 8.640 & \\
        \hline
        $R^2$ $\uparrow$ & \textbf{0.973} & 0.956 & 0.965\\
        \hline
        $R^2_{C_p}$ & 0.983 & 0.972 & 0.974 \\
        $R^2_{Cf_{x}}$ & 0.966 & 0.944 & 0.959 \\
        $R^2_{Cf_{y}}$ & 0.969 & 0.951 & 0.960 \\
        $R^2_{Cf_{z}}$ & 0.973 & 0.957 & 0.965\\
        \hline
        $\mathrm{wrMAE}$ $\downarrow$ & \textbf{0.210} & 0.228 & \\
        \hline
        $\mathrm{wrMAE}_{C_p}$ & 0.157 & 0.198 & \\
        $\mathrm{wrMAE}_{Cf_{x}}$ & 0.170 & 0.163 & \\
        $\mathrm{wrMAE}_{Cf_{y}}$ & 0.315 & 0.314 & \\
        $\mathrm{wrMAE}_{Cf_{z}}$ & 0.197 & 0.237 & \\
        \hline
        \# parameters $\downarrow$ & $\boldsymbol{7.3\times10^7}$ & $1.7\times10^{10}$ & $8.13\times10^7$ \\
        \hline
    \end{tabular}
\end{table*}

\subsection{Qualitative Flow Predictions}
\begin{figure*}[t]
    \centering
    \begin{subfigure}[t]{0.48\textwidth}
        \centering
        \includegraphics[width=\linewidth]{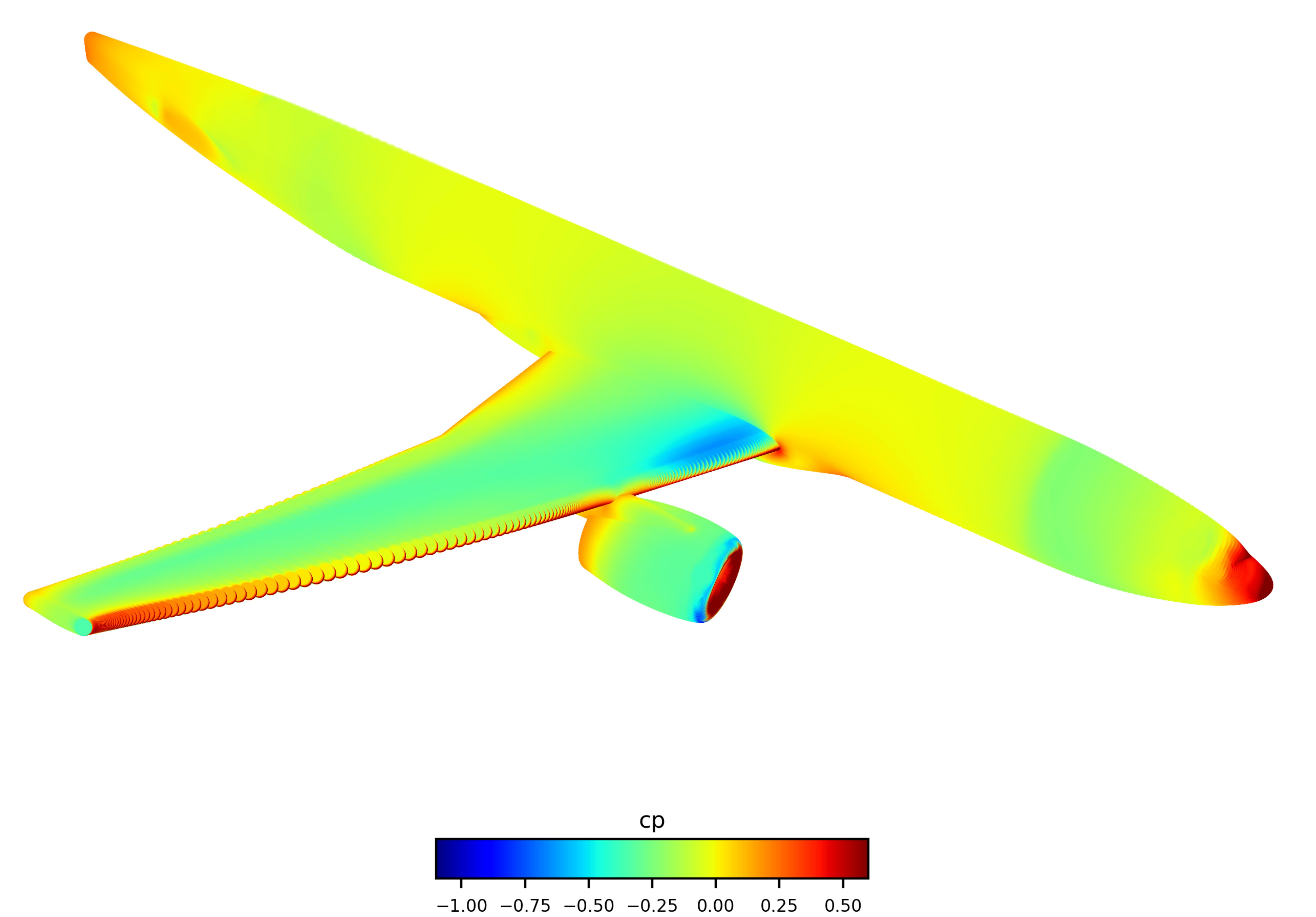}\\[-0.2em]
        \includegraphics[width=\linewidth]{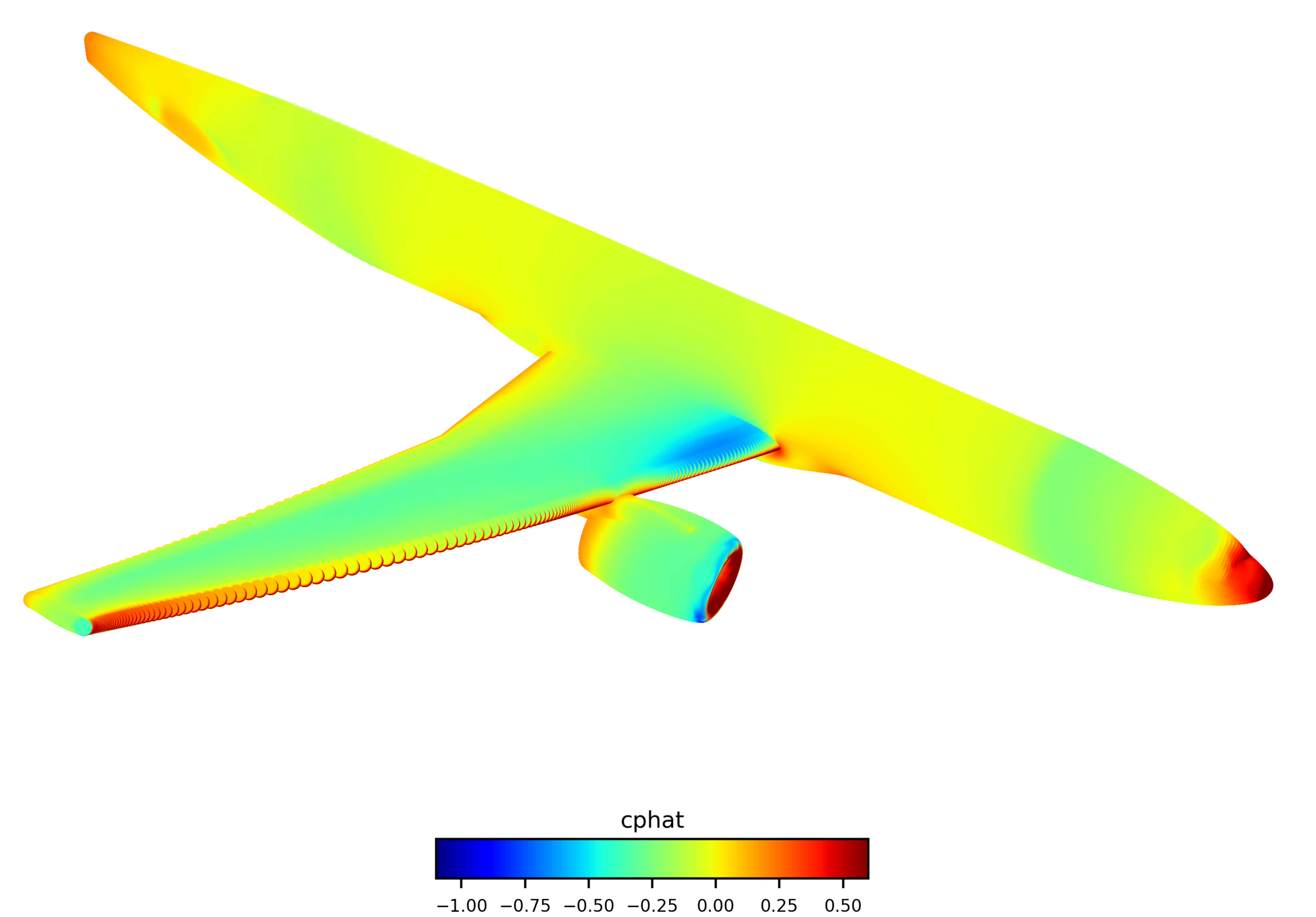}\\[-0.2em]
        \includegraphics[width=\linewidth]{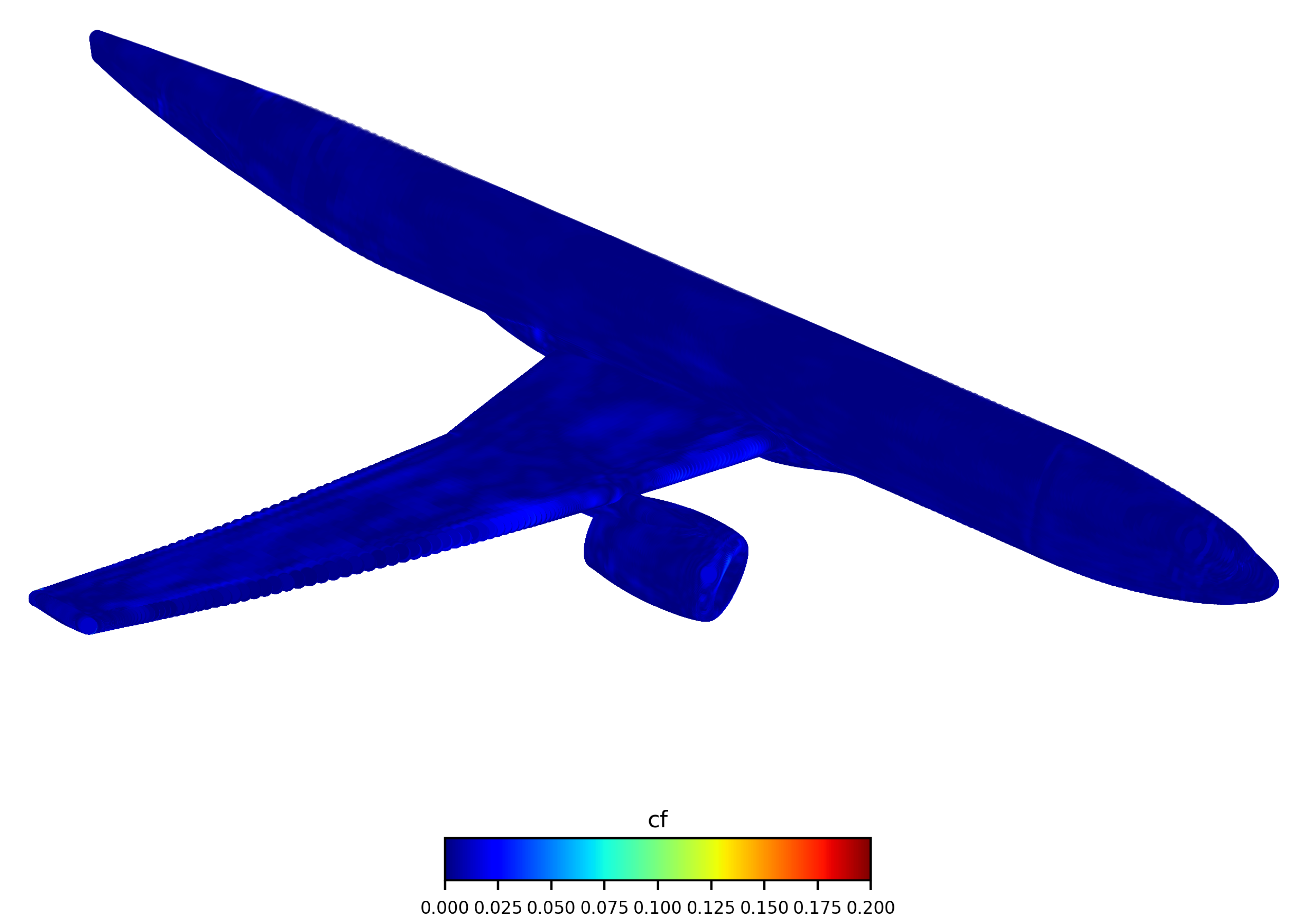}
        \caption{Pressure coefficient $C_p$. From top to bottom: CFD reference, model prediction, and absolute prediction error.}
        \label{fig:qualitative_results_1:cp}
    \end{subfigure}
    \hfill
    \begin{subfigure}[t]{0.48\textwidth}
        \centering
        \includegraphics[width=\linewidth]{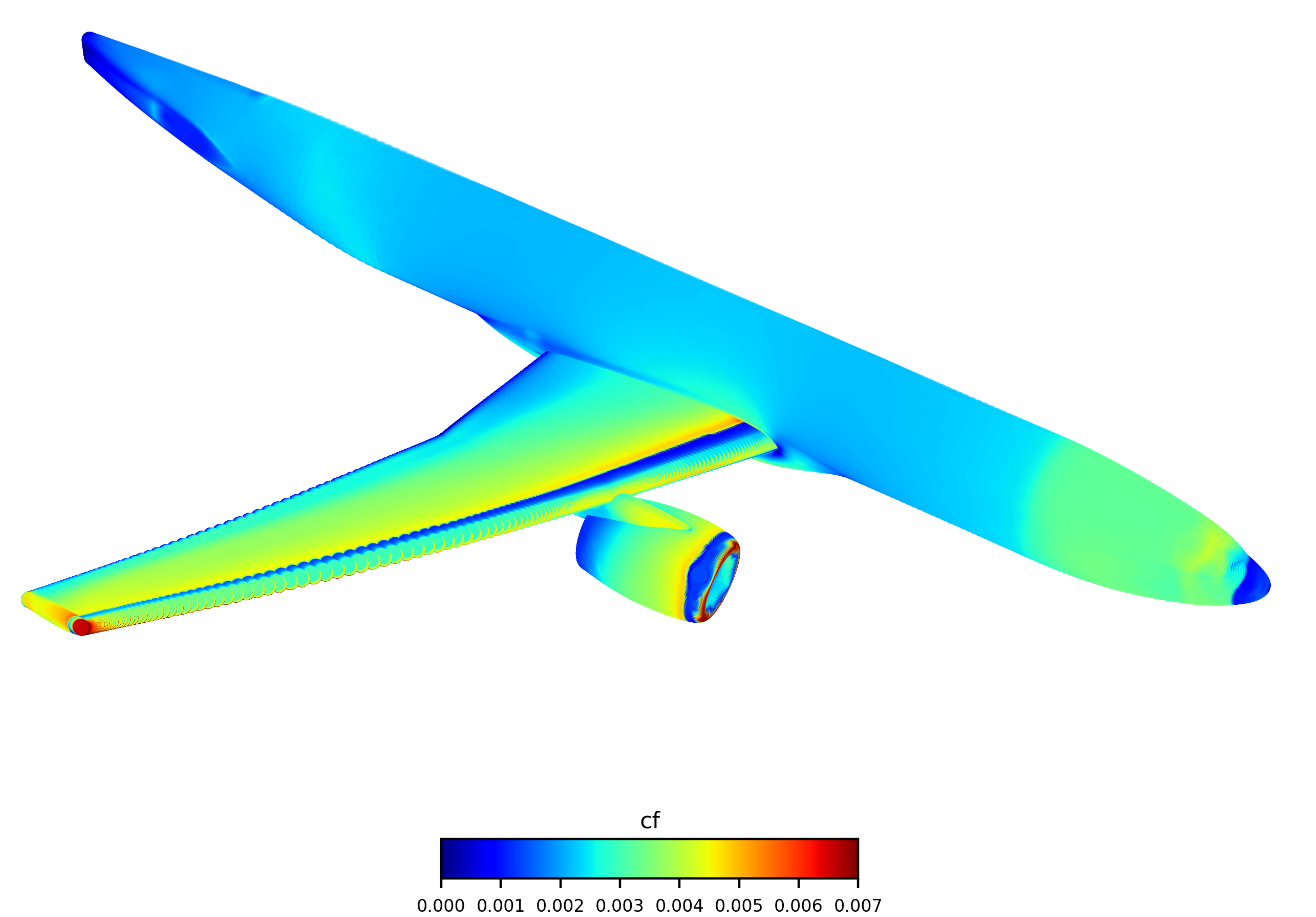}\\[-0.2em]
        \includegraphics[width=\linewidth]{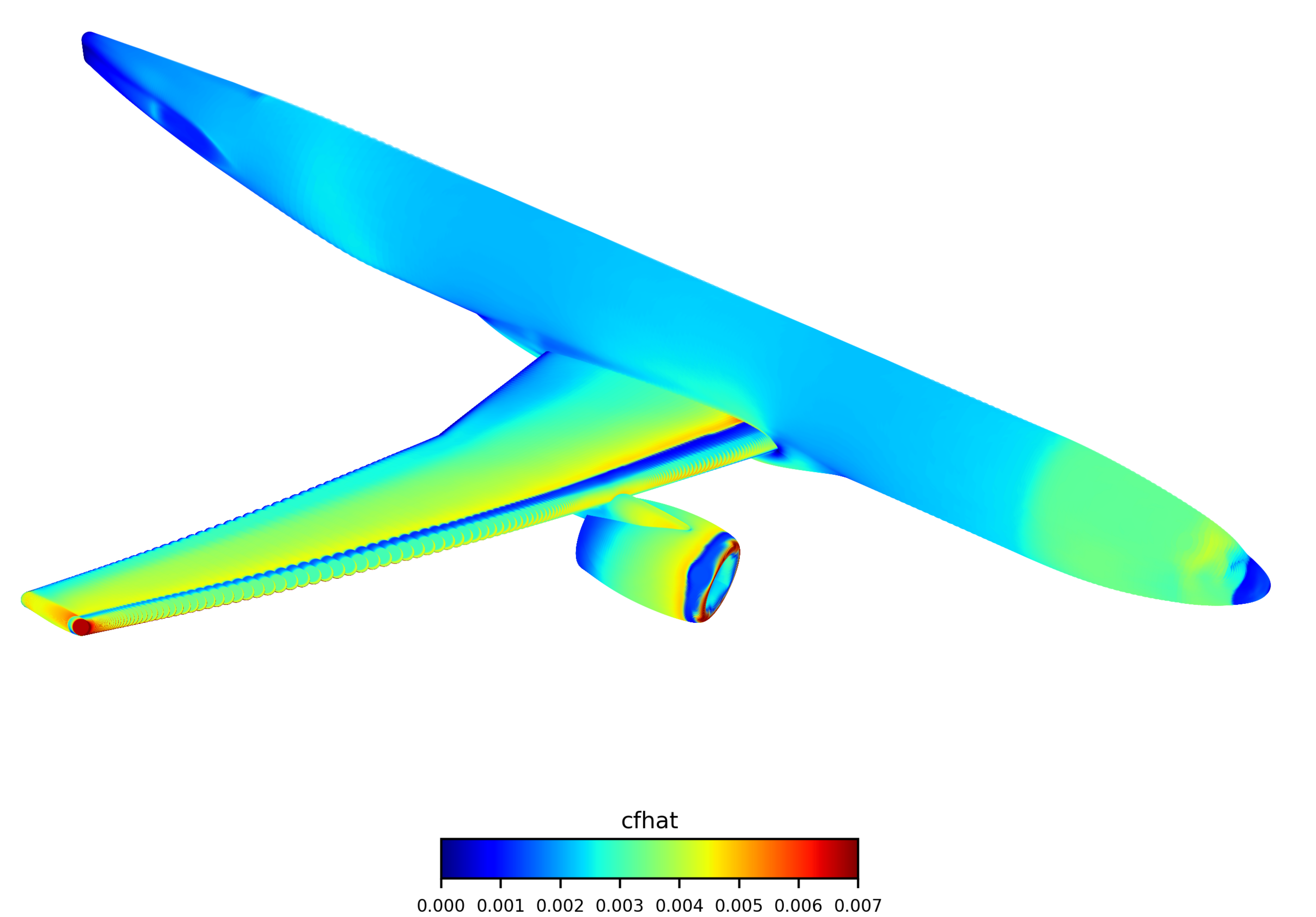}\\[-0.2em]
        \includegraphics[width=\linewidth]{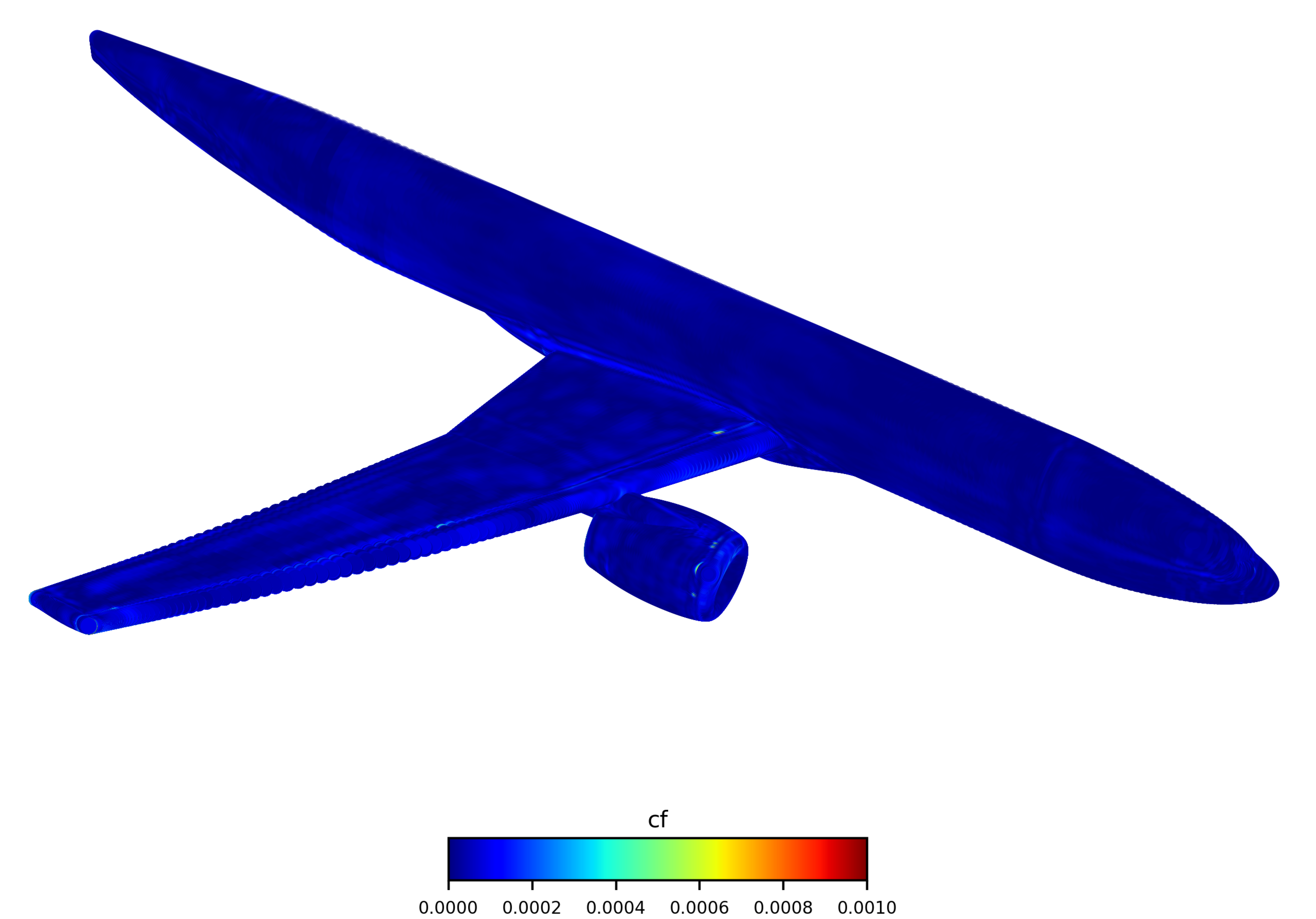}
        \caption{Norm of the friction field $\left(Cf_x, Cf_y, Cf_z\right)$. From top to bottom: CFD reference, model prediction, and absolute prediction error.}
        \label{fig:qualitative_results_1:cf}
    \end{subfigure}
    \caption{Pressure and friction fields prediction obtained with the proposed methodology for the aerodynamic conditions $M_\infty = 0.5$, $AoA = 0.0^\circ$ and $p_i = 4.0 \times 10^5$ Pa.}
    \label{fig:qualitative_results_1}
\end{figure*}

\begin{figure*}[!htp]
    \centering
    \begin{subfigure}[t]{0.48\textwidth}
        \centering
        \includegraphics[width=\linewidth]{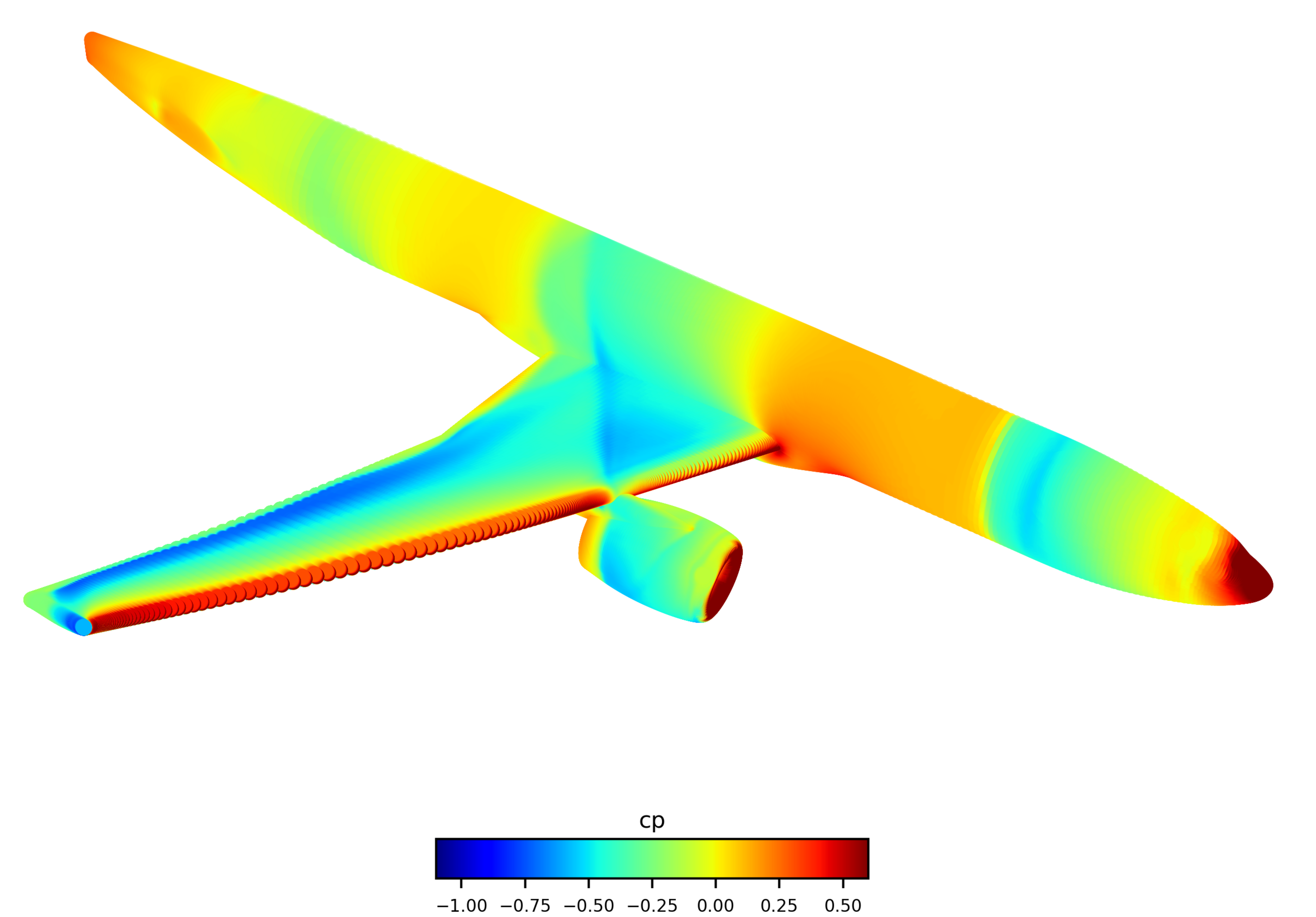}\\[-0.2em]
        \includegraphics[width=\linewidth]{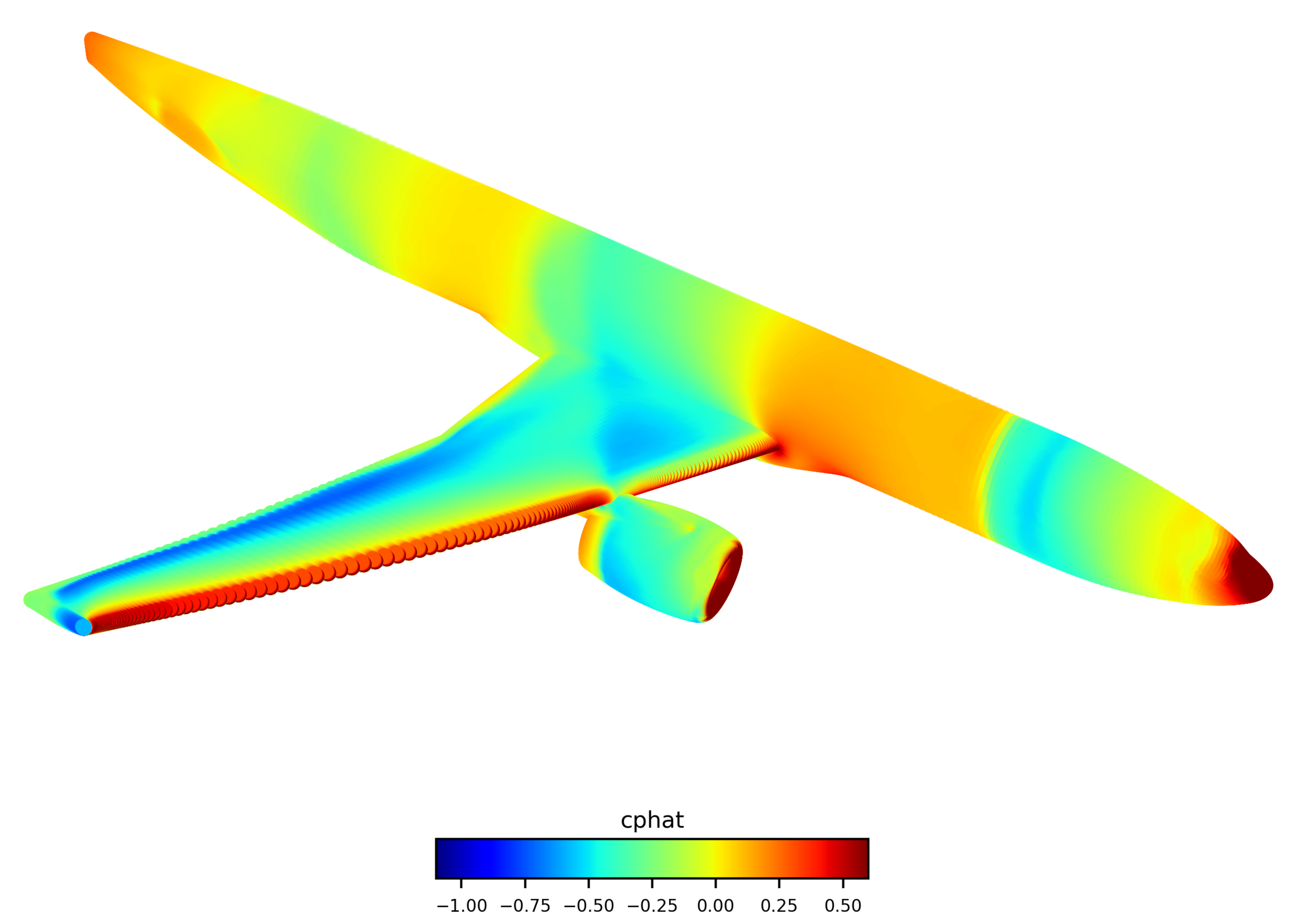}\\[-0.2em]
        \includegraphics[width=\linewidth]{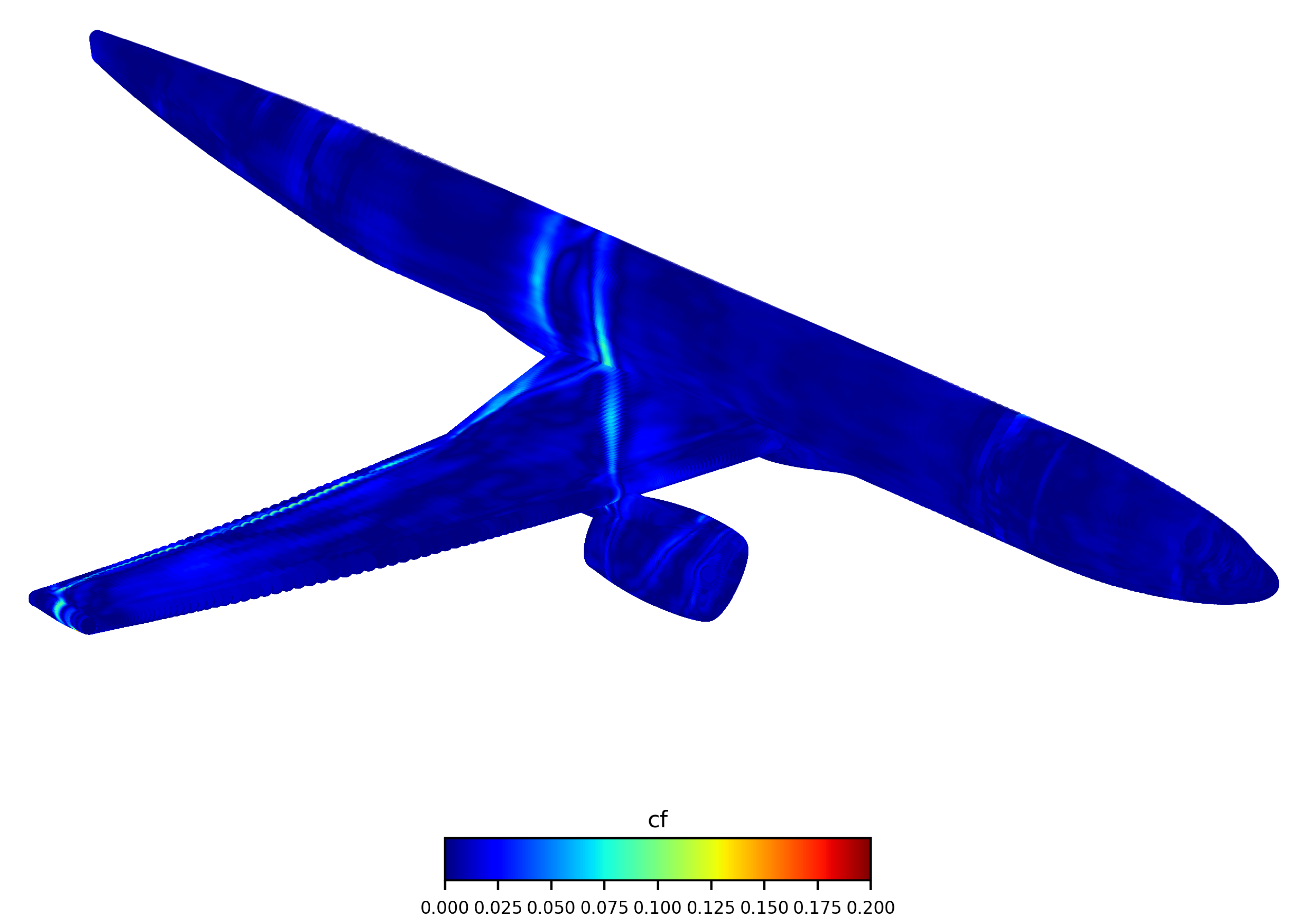}
        \caption{Pressure coefficient $C_p$. From top to bottom: CFD reference, model prediction, and absolute prediction error.}
        \label{fig:qualitative_results_4:cp}
    \end{subfigure}
    \hfill
    \begin{subfigure}[t]{0.48\textwidth}
        \centering
        \includegraphics[width=\linewidth]{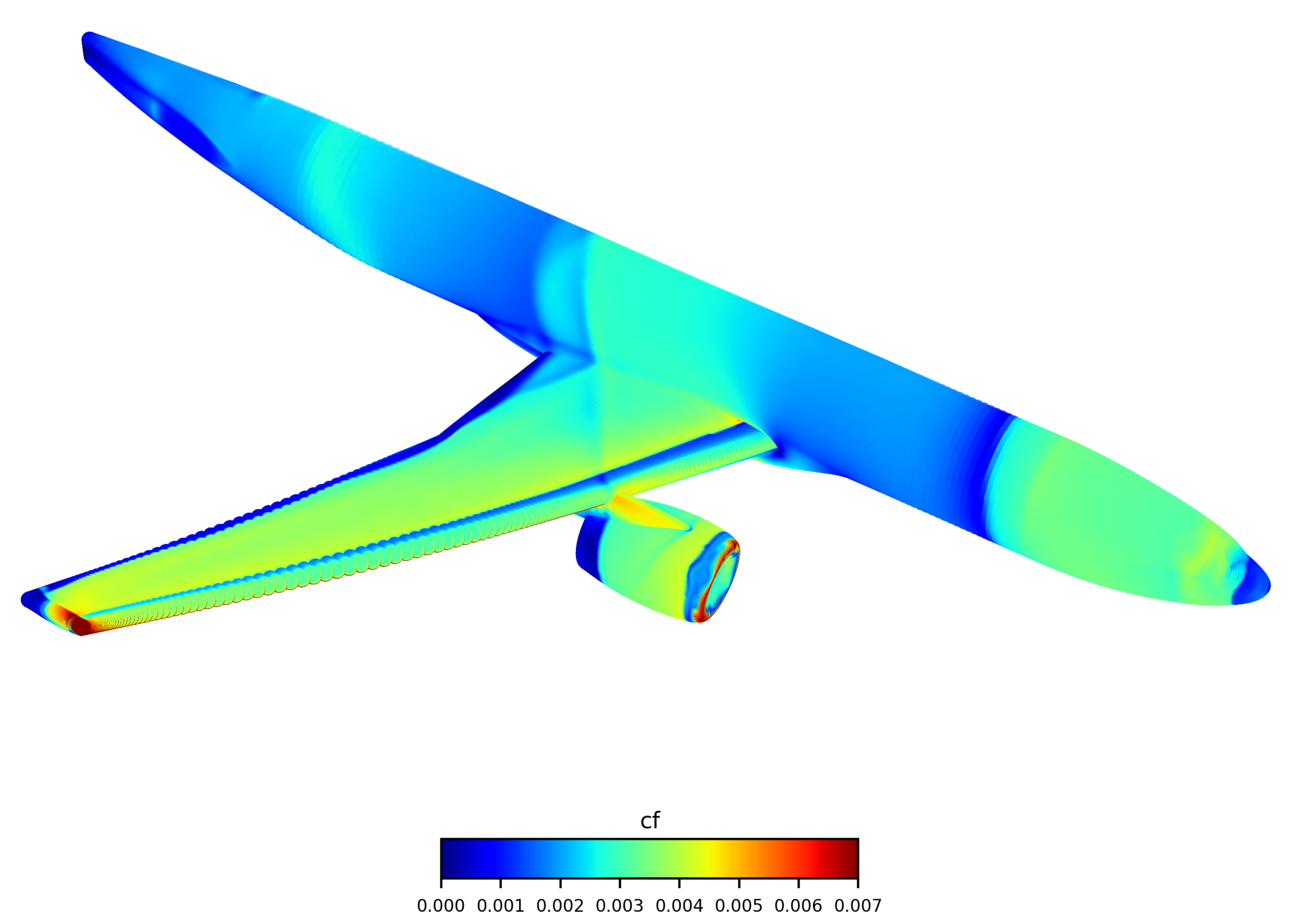}\\[-0.2em]
        \includegraphics[width=\linewidth]{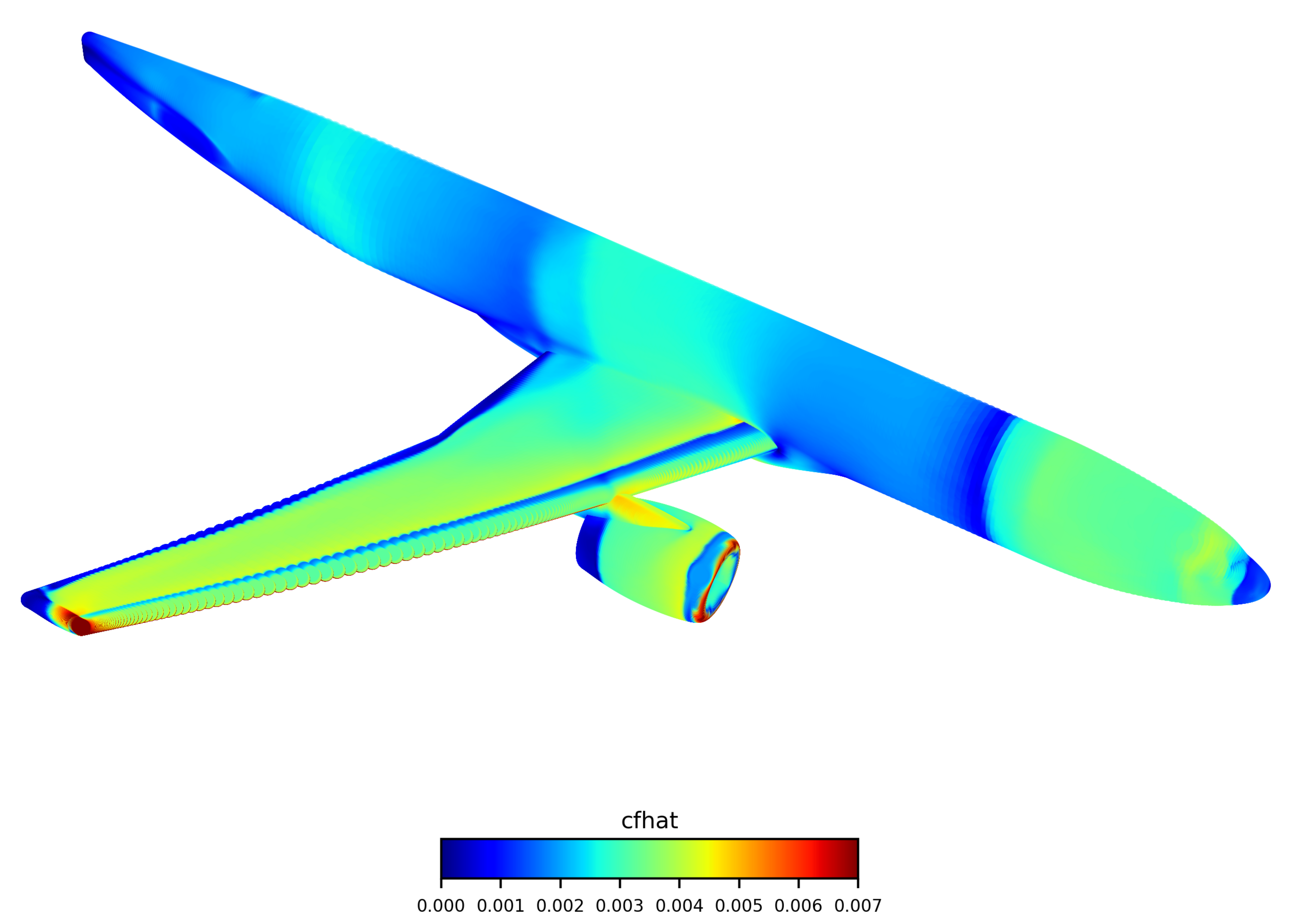}\\[-0.2em]
        \includegraphics[width=\linewidth]{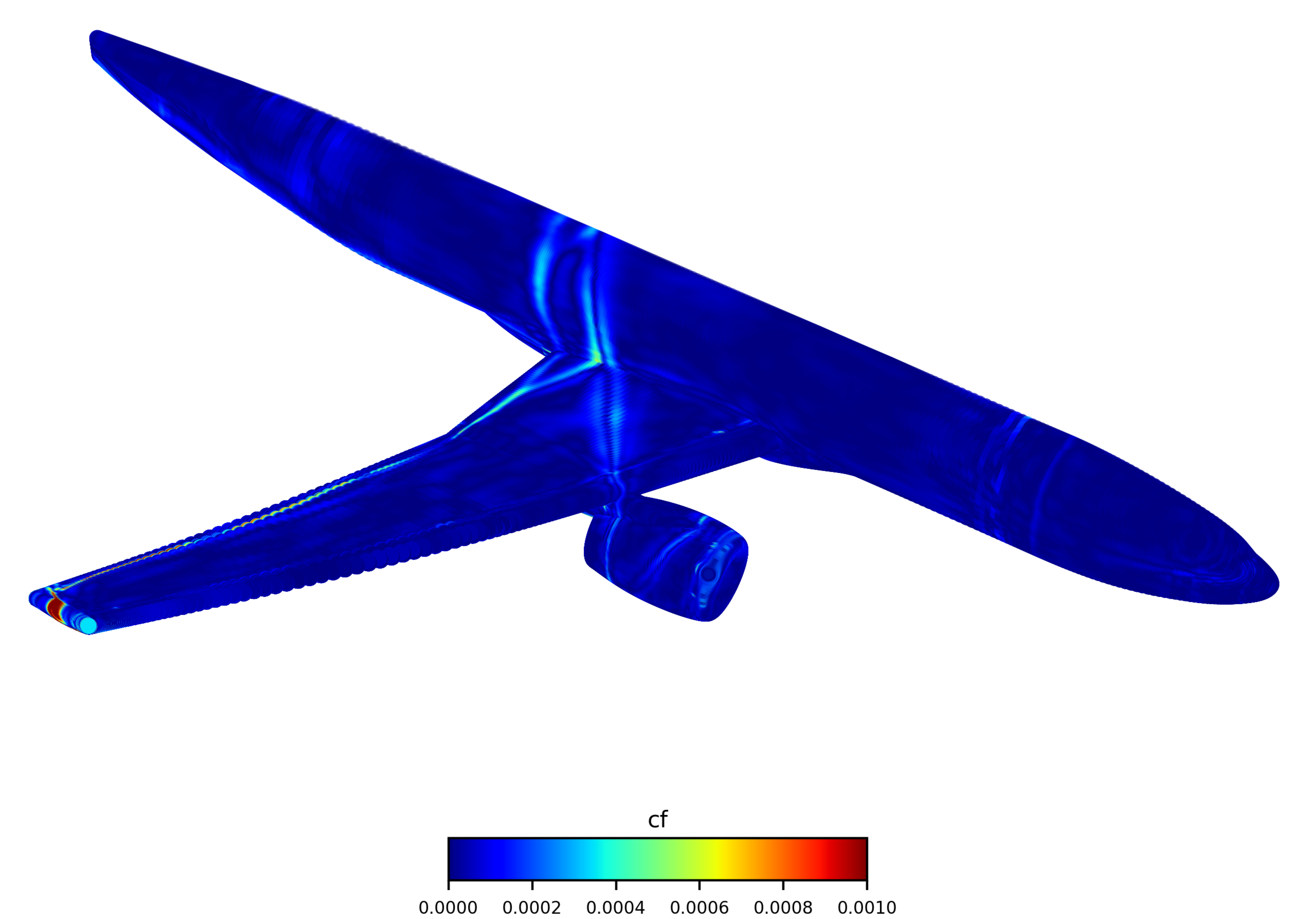}
        \caption{Norm of the friction field $\left(Cf_x, Cf_y, Cf_z\right)$. From top to bottom: CFD reference, model prediction, and absolute prediction error.}
        \label{fig:qualitative_results_4:cf}
    \end{subfigure}
    \caption{Pressure and friction fields prediction obtained with the proposed methodology for the aerodynamic conditions $M_\infty = 0.96$, $AoA = -2.0^\circ$ and $p_i = 2.0 \times 10^5$ Pa.}
    \label{fig:qualitative_results_4}
\end{figure*}

Beyond the quantitative evaluation, visual inspection of the predicted flow fields provides additional insight into the behavior of the surrogate model. Figures~\ref{fig:qualitative_results_1} and \ref{fig:qualitative_results_4} compares representative predictions obtained with the proposed methodology against the reference CFD solutions for several operating conditions including high Mach and high pressure regimes.


Overall, the predicted aerodynamic fields remain in close agreement with the CFD reference solutions over the entire aircraft surface. The largest discrepancies are generally confined to localized regions exhibiting strong spatial gradients, such as transonic shocks or areas affected by flow separation. 

\subsection{Comparison with recent surrogate modeling approaches} \label{sec:ff_results}
In addition to the organizer-provided baseline, the proposed methodology can be compared with more recent approaches developed for the same aerodynamic benchmark. In particular, Ramos et al.~\cite{ramos2026fluidflow} introduced FluidFlow, a generative surrogate model based on conditional flow matching and diffusion transformers, which was independently evaluated on the ONERA CRM-WBPN dataset using the official training and test split.

Their approach operates directly on the unstructured surface representation and achieves a global $R^2$ value of 0.965, outperforming the organizer-provided Global MLP baseline (0.956). A comparison with the proposed neural-field ensemble reveals consistently higher predictive accuracy for the latter, which achieves an overall $R^2$ of 0.973 (see Table \ref{tab:final_results}). Improvements are consistently observed across all four predicted aerodynamic quantities, including both the pressure coefficient and the three skin-friction components. However, Ramos et al.~\cite{ramos2026fluidflow} do not report the $\mathrm{wrMAE}$ metric, which is required to compute the official challenge score. Consequently, a direct comparison using the competition scoring formula is not possible based on their published results.

These results demonstrate the competitiveness of coordinate-based neural fields against more recent generative modeling approaches for aerodynamic surface prediction. In particular, they suggest that a relatively simple deterministic architecture can provide highly accurate predictions without requiring a more sophisticated generative framework.

\section{Conclusion} \label{section:conclusion}


This paper presented the methodology that achieved first place in the ONERA CRM Wall Distribution Regression Challenge for aerodynamic wall-field prediction on the CRM-WBPN benchmark \cite{peter2025crm}. Rather than introducing a fundamentally new neural architecture, the objective was to demonstrate how a careful combination of existing machine-learning techniques, guided by the characteristics of the benchmark and a systematic experimental analysis, can substantially improve predictive performance.

The proposed surrogate model formulates aerodynamic field prediction as a conditional neural field operating directly on point coordinates, surface normals, and aerodynamic operating conditions. Starting from a simple multilayer perceptron prototype, the final methodology was progressively developed through the introduction of FFE, an increased model capacity, a RSE loss aligned with the evaluation metric, an ensemble learning strategy, and $k$-fold cross-validation. The progressive ablation study shows that each of these components contributes to the final performance, while the investigation of several alternative strategies provides additional insight into the design of surrogate models for complex aerodynamic configurations.

On the hidden test set, the final methodology achieved the highest score in the competition, outperforming the strongest organizer-provided baseline across both global accuracy and worst-case prediction metrics. Although the proposed solution relies on an ensemble of neural fields, it remains considerably more parameter-efficient than the Global MLP baseline while providing more accurate predictions of localized flow structures, including transonic shocks and sharp variations in the skin-friction fields. These results demonstrate that coordinate-based neural fields, combined with appropriate training strategies, constitute an interesting framework for aerodynamic surrogate modeling when mesh connectivity is unavailable and the amount of training data is limited.

Furthermore, comparison with FluidFlow, a more recent generative approach independently evaluated on the same CRM-WBPN benchmark, shows that the proposed neural-field ensemble achieves consistently higher $R^2$ values across all four predicted aerodynamic quantities.

\paragraph{Perspectives.}
Several directions could further improve the proposed methodology and broaden its applicability. A first perspective consists in studying the influence of the training database size on the model prediction accuracy in order to better characterize the data requirements of neural-field surrogate models. In addition, active learning strategies could be explored to selectively identify the most informative training samples and maximize performance gains. Evaluating the practical deployment of such surrogates in industrial workflows also deserves further investigation, particularly regarding the trade-off between model size, inference time, and predictive accuracy.

Another important direction is to extend the methodology beyond a single aircraft configuration by considering multiple geometries during training. This naturally leads to the development of foundation models for aerodynamic surrogate modeling, capable of learning generalized geometric and physical representations that could subsequently be adapted or fine-tuned to downstream tasks involving new aircraft configurations for which only sparse CFD data are available.

Future work could also investigate more advanced ensemble strategies and uncertainty quantification techniques in order to improve the reliability of the predictions and provide confidence estimates for engineering applications. Likewise, incorporating physics-informed weighting strategies during training, for example by emphasizing wing regions containing transonic shocks or other critical flow features, could further improve the prediction of localized discontinuities that dominate the overall error.

Finally, while the absence of mesh connectivity motivated the coordinate-based formulation adopted in this work, future benchmarks providing access to the computational mesh would enable direct comparisons with graph neural networks and mesh-based neural operators. Similarly, evaluating point-cloud architectures, such as Point Transformer models, would provide further insight into the respective advantages and limitations of coordinate-based, point-cloud, and topology-aware approaches for high-fidelity aerodynamic surrogate modeling.

\section*{Acknowledgments}
The authors would like to thank Jacques Peter and Quentin Bennehard (ONERA) for organizing the CRM Wall Distribution Regression Challenge and for simulating and releasing the CRM-WBPN benchmark dataset. Their efforts in designing a realistic and challenging benchmark, together with the public release of the dataset following the competition, provide a valuable resource for the development and evaluation of machine-learning methods for aerodynamic surrogate modeling.

This work is supported by the "ARIAC by DigitalWallonia4.ai" research project (grant agreement No 2010235 – TRAIL institute) and benefited from computational resources made available on Lucia, the Tier-1 supercomputer of the Walloon Region, infrastructure funded by the Walloon Region (grant agreement No 1910247).

\section*{AI Disclosure Statement}
Generative AI tools, such as ChatGPT, were used for language editing and improving the clarity of the manuscript. All intellectual content, research design, and data analysis were conducted solely by the authors.
\bibliographystyle{unsrt}  
\bibliography{references}  
\end{document}